\documentclass[sigconf,nonacm]{acmart}
\usepackage{pifont}
\usepackage{booktabs}
\usepackage{graphicx}
\usepackage{multirow}
\usepackage{colortbl} 
\usepackage{adjustbox}
\usepackage{array}
\usepackage{color}
\usepackage{xcolor}
\usepackage{caption}
\usepackage{xspace}
\usepackage{enumitem}
\usepackage{graphicx} 
\usepackage{tabularx} 
\usepackage{lipsum}  
\usepackage{makecell}
\newcommand{\fix}[1]{}

\newcommand{\eg}{\emph{e.g.},\xspace}

\newcommand{\etc}{\emph{etc.}\xspace}

\newcommand{\name}{\textsc{LLMAD}\xspace}

\usepackage[normalem]{ulem}
\useunder{\uline}{\ul}{}
\begin{document}

\title{Large Language Models can Deliver Accurate and Interpretable Time Series Anomaly Detection}

\author{Jun Liu}
\affiliation{%
  \institution{University of Chinese Academy of Sciences}
  \country{China}
}\authornote{This work was completed during their internship at Microsoft.}

\author{Chaoyun Zhang}
\affiliation{%
  \institution{Microsoft}
  \country{China}
}

\author{Jiaxu Qian}
\affiliation{%
  \institution{Zhejiang University of Technology}
  \country{China}
}\authornotemark[1]

\author{Minghua Ma}
\affiliation{%
  \institution{Microsoft}
  \country{USA}
}

\author{Si Qin}
\affiliation{%
  \institution{Microsoft}
  \country{China}
}

\author{Chetan Bansal}
\affiliation{%
  \institution{Microsoft}
  \country{USA}
}

\author{Qingwei Lin}
\affiliation{%
  \institution{Microsoft}
  \country{China}
}

\author{Saravan Rajmohan}
\affiliation{%
  \institution{Microsoft}
  \country{USA}
}

\author{Dongmei Zhang}
\affiliation{%
  \institution{Microsoft}
  \country{China}
}

\renewcommand{\shortauthors}{Liu et al.}

\begin{abstract}
    Time series anomaly detection (TSAD) plays a crucial role in various industries by identifying atypical patterns that deviate from standard trends, thereby maintaining system integrity and enabling prompt response measures. Traditional TSAD models, which often rely on deep learning, require extensive training data and operate as black boxes, lacking interpretability for detected anomalies. To address these challenges, we propose \textbf{\name}, a novel TSAD method that employs Large Language Models (LLMs) to deliver accurate and interpretable TSAD results. \name innovatively applies LLMs for few-shot anomaly detection by retrieving and leveraging both positive and negative similar time series segments, significantly enhancing LLMs' effectiveness. Furthermore, \name employs the Anomaly Detection Chain-of-Thought (AnoCoT) approach to mimic expert logic for its decision-making process. This method further enhances its performance and enables \name to provide explanations for their detections through versatile, customized perspectives, which are particularly important for user decision-making. Experiments on three datasets indicate that our \name achieves detection performance comparable to state-of-the-art deep learning methods while offering remarkable interpretability for detections. To the best of our knowledge, this is the first work that directly employs LLMs for TSAD.
\end{abstract}






\maketitle


\section{Introduction\label{sec:intro}}

\begin{figure}[t]
\centering
\includegraphics[width=\columnwidth]{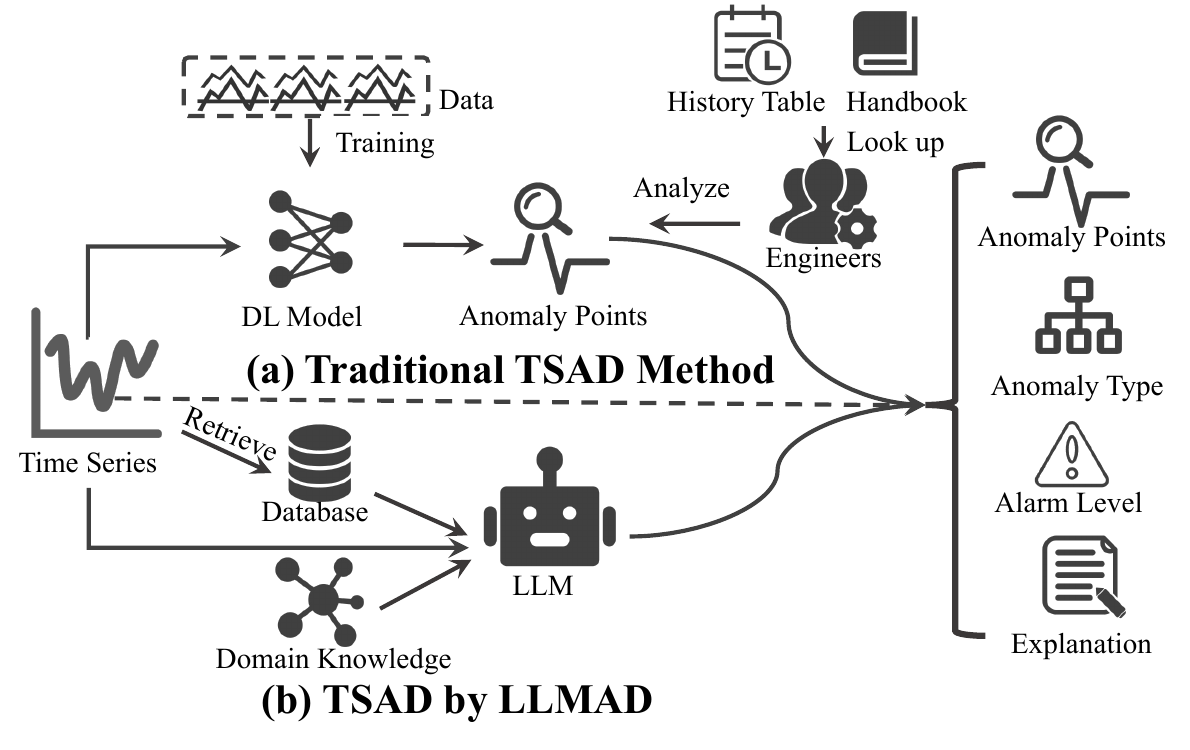}
\caption{Time series anomaly detection for traditional deep learning methods and LLMs.}
\label{fig:compare}
\end{figure}
In the data-driven landscape of contemporary industries, time series data holds a pivotal role across various sectors, including financial market fluctuations \cite{ahmed2017anomaly}, manufacturing production line monitoring \cite{hsieh2019unsupervised}, and internet traffic analysis \cite{cook2019anomaly}. The examination and understanding of time series data are invaluable for maintaining operational stability, mitigating potential financial risks, and ensuring cybersecurity. Time Series Anomaly Detection (TSAD) is a critical task within these applications, aimed at identifying atypical points or patterns in the data stream that may lead to system malfunctions, market disruptions, or security threats, \cite{blazquez2021review, schmidl2022anomaly,  jin2023assess, chen2023imdiffusion} \etc. By providing alerting signals, TSAD contributes to maintaining operational efficiency and minimizing disruptions \cite{schmidl2022anomaly, jacob2021exathlon}. This process is particularly facilitated when the interpretation of anomaly information is readily available \cite{rad2021explainable, tang2023gru}.

Deep learning-based anomaly detectors have recently gained prominence in the TSAD domain, primarily due to their high detection accuracy, which ensures reliability \cite{blazquez2021review, zeng2023traceark, ma2018robust, zhang2019deep}. However, achieving this level of performance necessitates massive training data for learning, requiring significant data collection efforts. Moreover, such detectors typically only provide detected anomalous points, functioning as black boxes and failing to offer additional interpretation, such as anomaly type, emergency level, and textual anomaly reports in natural language. Such information is however particularly crucial for understanding the underlying reasons behind detected anomalies and for gaining the trust of domain experts who rely on these models for decision-making \cite{jacob2020exathlon, amarasinghe2018toward}. Generating these human-readable reports often demands substantial manual analysis, involving consultation of historical data and handbooks, which is both time-consuming and requires specialized knowledge, as illustrated in Figure~\ref{fig:compare} (a). Consequently, the usability of deep learning anomaly detectors in practical applications is limited. 
\begin{sloppypar}
The remarkable achievements of Large Language Models (LLMs)~\cite{achiam2023gpt} in various domains~\cite{zhang2024ufo, zhang2024allhands, ding2023everything} have led to their application in time series tasks, garnering significant attention~\cite{gruver2024large, zhou2024one}. LLMs are pretrained with diverse textual data, including numerical time series \cite{jin2024position}, and excel in in-context learning (ICL) with only a few-shot examples \cite{min2022rethinking, jin2023assess, chen2024automatic}. They are also adept at providing more interpretable, human-readable explanations for their predictions and can be customized for specific domains. These attributes make LLMs particularly suitable for the TSAD domain, where high interpretability is highly desirable. As illustrated in Figure\ref{fig:compare}, deep learning-based detectors (as shown in {Figure\ref{fig:compare}~}(a)) necessitate considerable human effort to obtain interpretable anomaly reports, whereas LLMs (as shown in {Figure\ref{fig:compare}~}(b)) can directly output the required information alongside their predictions without human intervention. LLMs can provide anomaly points, anomaly types, alarm levels, and explanations by integrating relevant knowledge directly into the prompts, such as data background information and anomaly detection rules. This integration allows predictions to align more closely with domain-specific requirements for TSAD.
\end{sloppypar}
Applying LLMs directly to TSAD poses several challenges. Anomaly detection requires an in-depth understanding and background knowledge of the data in order to distinguish between normal fluctuations and genuine anomalies. Achieving this distinction becomes difficult if  decision boundaries are not properly provided or if domain knowledge is insufficient. Moreover, the quality of interpretability is intimately connected to the method's comprehension of the data, as a more profound understanding of the data results in more accurate explanations that are consistent with the data's background and patterns. This calls for the integration of domain knowledge from human expertise into LLMs to augment their understanding of the data.

To harness the power of LLMs while addressing these challenges, we introduce \name, an LLM-based anomaly detection framework specifically tailored for TSAD, offering high accuracy and interpretability. \name builds upon existing pretrained LLMs to generate anomaly predictions and explanations without the need for fine-tuning with time series data. To enable \name to acquire better knowledge of the data, it retrieves both normal and abnormal patterns from history as input to activate the In-Context Learning (ICL)~\cite{brown2020language}, thereby enabling accurate anomaly prediction with few shots. Furthermore, \name employs Anomaly Detection Chain of Thoughts (AnoCoT) prompting \cite{wei2022chain} to  incorporate domain knowledge tailored to TSAD, which improves its prediction performance while providing more logical and human-readable interpretations. With these techniques, \name can not only accurately detect anomaly points but also offer explanations for its reasoning process and classifications of key components, such as anomaly type and alarm level, providing valuable insights for decision-making.

We conduct experiments on three mainstream public datasets for the task of TSAD \cite{li2022constructing, laptev2015benchmark, zhang2022efficient}. The results suggest that our \name achieves comparable accuracy performance with state-of-the-art (SOTA) deep learning-based detectors. Importantly, \name can further deliver useful and readable interpretable results, including anomaly type, alarm level classification, and explanation, at a low cost, as evaluated by domain experts. These outcomes cannot be achieved by traditional methods. Overall, this paper makes the following contributions:

\begin{enumerate}[leftmargin=*]
    \item We introduce \name, a novel framework based on LLMs, which accurately conducts point-wise TSAD while providing comprehensive interpretation of its predictions to facilitate decision-making.
    \item We inject data background and domain knowledge into \name via time-series In-Context Learning (ICL) and AnoCoT to enhance TSAD performance and interpretation quality.
    \item We conduct extensive experiments on three public datasets. Both quantitative and human evaluations suggest that \name delivers highly accurate TSAD while providing useful and readable interpretative results at a low cost.
\end{enumerate}
To the best of our knowledge, we are the first to employ LLMs to directly perform TSAD without fine-tuning, delivering interpretable results.

\section{Preliminary}
In this section, we provide an overview of TSAD and its interpretability.

\subsection{Time Series Anomaly Detection}
Anomaly detection, also referred to as outlier detection, is an analytical process aimed at identifying a single point or a sequence of points that exhibit significant deviations from the typical patterns or norms~\cite{dang2020time}. 
A univariate time series is defined as:
\begin{equation}
T=\left\{x_i \mid i \in[1, n]\right\}
\end{equation}
where $x_i$ denotes the data value at time $i$, within the interval $[1, n]$. The objective of anomaly detection is to identify instances of abnormal behavior (either normal or anomalous) at each time step. The result of an anomaly detection algorithm is represented as:
\begin{equation}
Y=\left\{y_i \mid i \in[1, n]\right\}
\end{equation}
where $y_i \in\{0,1\}$ serves as a binary indicator, with 0 indicating normality and 1 signifying abnormality.

\begin{figure*}[t]
\centering
\includegraphics[width=1\textwidth]{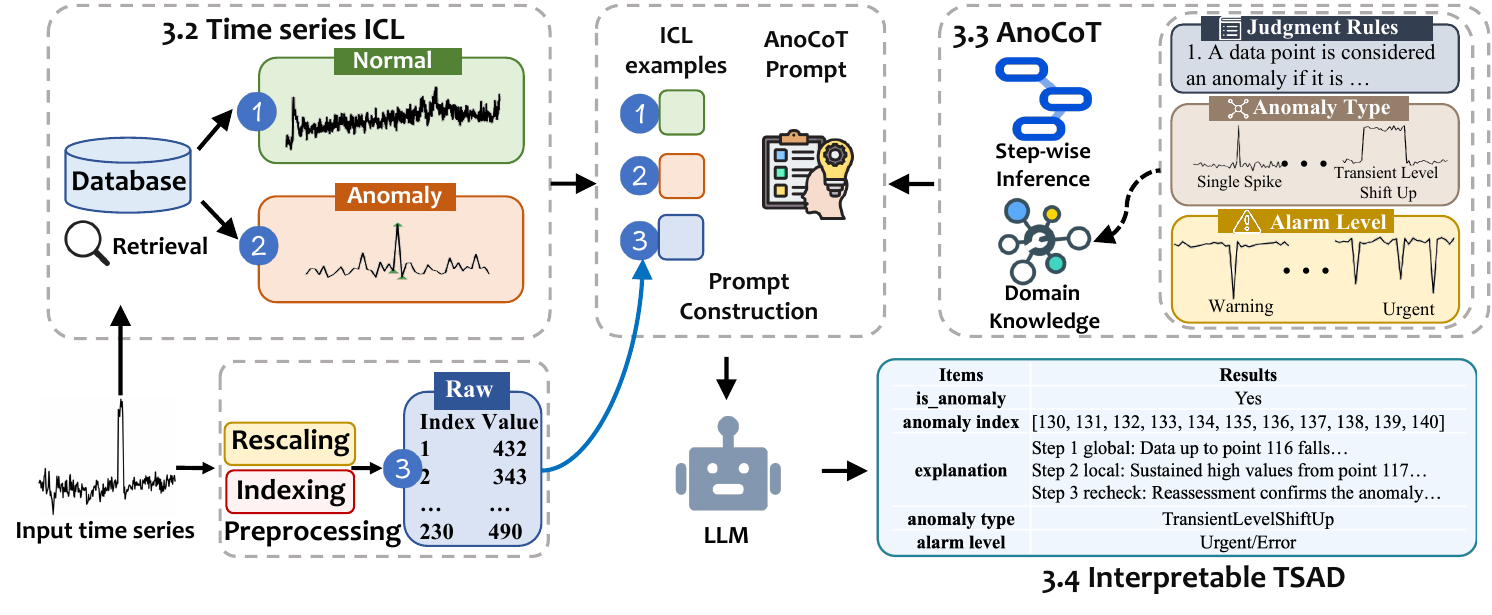}
\caption{The overview of \name.}
\label{fig:generation_pic}
\end{figure*}

\subsection{Interpretability for Time Series Anomaly Detection} 
Interpretability in TSAD is crucial for transforming raw data insights into actionable intelligence~\cite{jacob2020exathlon}. An interpretable TSAD system not only identifies anomalies but also provides explanations that can be understood by humans, facilitating informed decision-making and root cause analysis \cite{chen2024automatic, wang2023root}. In this paper, we employ three key aspects of interpretability for TSAD, which are discussed below: text explanations~\cite{yu2023temporal}, anomaly types and alarm levels~\cite{jacob2020exathlon}. \fix{citation}

\noindent\textbf{(1) Text Explanations:} Text explanations involve providing human-readable descriptions that clarify why a particular data point or sequence is considered anomalous. These explanations are essential for users who may not have deep technical expertise but need to understand the nature of an anomaly to respond appropriately.

\noindent\textbf{(2) Anomaly Types:} Classifying anomalies into different types (\eg spike, shift up) enhances the interpretability of TSAD by categorizing unusual patterns based on their characteristics and potential implications. This classification helps in prioritizing responses and understanding the anomaly's context within the broader system behavior.

\noindent\textbf{(3) Alarm Levels:} Alarm levels indicate the severity of detected anomalies, guiding immediate attention and resources to the most critical issues. By assigning a severity level to each detected anomaly, organizations can triage responses, allocate resources, and prioritize more effectively.

Leveraging the above information beyond anomaly prediction enables users to make better-informed decisions, take appropriate actions, and make reasonable judgments. This can be achieved automatically with \name, whereas it is challenging for traditional black-box methods.

\section{The Design of \name}
We provide an overview of our proposed \name in Section~\ref{sec:nutshell} and elaborate on its components in the subsequent subsections.

\subsection{\name in a Nutshell \label{sec:nutshell}}

We present an overview of \name in Figure~\ref{fig:generation_pic}. \name builds upon a time series database to retrieve similar examples for the model and employs AnoCoT prompting to inject domain knowledge into LLMs, helping it grasp the background of the time series data without the need for fine-tuning. The model then uses this information to predict anomaly points and provide interpretations.

Upon receiving the input time series, the data undergoes normalization and preprocessing to convert it into a tabular format suitable for LLM processing. \name then retrieves similar data samples, including both normal and abnormal data, from the time series databases into its inputs to help differentiate between the two data types (Section~\ref{sec:rag}). Then, it employs the AnoCoT to inject domain knowledge of anomaly rules, types, and alarm levels tailored to the specific dataset and constructs the prompts fed into the LLM (Section~\ref{sec:prompt}). With this information, \name predicts the anomaly points, along with their types, alarm levels, and textual explanations to provide comprehensive interpretations (Section~\ref{sec:interpretation}). We detail each component in the following subsections.

\subsection{Time Series Data Preprocessing}
Since LLMs are not primarily designed for numerical data, we conduct several data preprocessing steps on the time series to make them suitable for LLM input and improve the performance for TSAD. These steps include rescaling and indexing.

\textbf{Rescaling:}
To reduce tokens in LLM when the inputs are very large or high decimal places, we scale fractional values in integers with a certain number of significant digits \cite{gruver2024large}. We apply the following affine transformation to each element $x_t$ in a time series:
\begin{align}
x't = \left\lfloor 1000 \times \frac{(x_t - b)}{a} \right\rceil
\end{align}
Here, $\left\lfloor \cdot \right\rceil$ is the rounding to integral operator, and $a$ is the 95\textsuperscript{th} upper percentile of the data subset.
\begin{align}
b = \min{x_t} - \beta (\max_{x_t} - \min_{x_t}),
\end{align}
where $\min_{x_t}$ and $\max_{x_t}$ are the 1\textsuperscript{st} and 99\textsuperscript{th} percentiles, respectively, and $\beta$ is a scaling parameter. This scaling ensures that the values are integral and robust to outliers, making the input more suitable for LLMs in terms of understanding and token saving.







\textbf{Value Indexing:}
The primary goal of TSAD is to identify the exact data index of anomalous points. Therefore, we reformat the data values with numerical indices in a tabular format as follows:
\begin{align*}
    \text{Index} \:\:& \text{Value}\\
    1 \:\:\:\:\:& value 1\\
    2 \:\:\:\:\:& value 2\\
    \cdots \:\:\:& \cdots
\end{align*}
This correspondence assigns unique indices to each data value in the time series, thus facilitating LLMs to output point-wise anomalies \emph{with their indices as identifiers}, meeting the primary requirements of TSAD.

\subsection{Time Series In-Context Learning \label{sec:rag}}
The objective of an anomaly detection system is to delineate a distinct boundary between normal and abnormal data. Nevertheless, establishing such criteria poses a challenge due to the varying anomaly patterns across different datasets, necessitating substantial background and domain-specific knowledge. Deep learning-based detectors attempt to uncover this boundary by utilizing extensive training data, which results in a highly inefficient process. In contrast, LLMs have already undergone training with massive datasets, encompassing numerical time series, thereby requiring only a limited number of demonstrations to activate and apply their domain knowledge in the realm of TSAD. This methodology, referred to as In-Context Learning (ICL), has demonstrated its efficacy in numerous applications.

\begin{figure}[t]
\centering
\includegraphics[width=1\columnwidth]{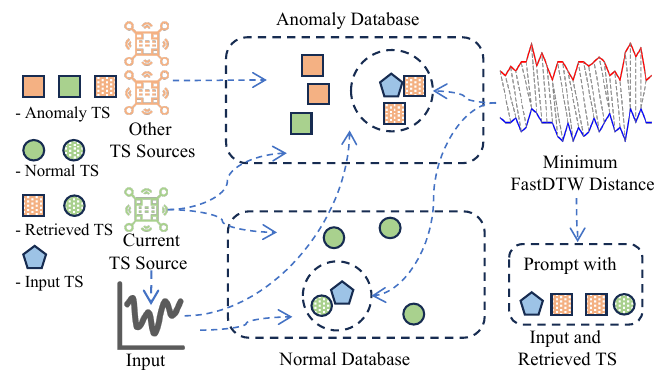}
\caption{Construction of Time Series Retrieval Databases and Time Series Retrieval.\fix{remove LLM}}
\label{fig:rag}
\end{figure}

We apply the similar ICL in \name to address the challenges in TSAD, where we build a time series database that store both normal and abnormal data for retrieval. When receiving an input time series, we find both top-$k$ similar normal and abnormal samples in the database, to establish a boundary between normal and abnormal data with them. We illustrate this process in Figure~\ref{fig:rag}.

\textbf{Construction and Retrieval of Databases.} We first collect both normal and abnormal time series data and retrieve examples from them individually. The normal time series database, denoted as $\tilde{\mathcal{S}}$, comprises normal time series derived from the same dataset. In contrast, the anomaly database, denoted as $\hat{\mathcal{S}}$, is assembled from all abnormal time series sources across different subsets within the dataset to expand the variety and quantity of anomaly types, due to the rareness of abnormal data. 

\textbf{Finding Similar Time Series.} Identifying similar time series differs from the methods employed for textual data, where fixed-length vectors are embedded and cosine similarity is utilized as a measurement~\cite{zamani2016embedding}. For time series analysis, methods like embedding combined with cosine similarity are less favorable because they often require training data to generate meaningful embeddings and might not robustly account for temporal shifts and distortions between sequences~\cite{wang2013experimental}.  In contrast, Dynamic Time Warping (DTW)~\cite{keogh2005exact} and its optimized version FastDTW~\cite{salvador2004fastdtw}, directly address these issues by explicitly aligning sequences by matching temporal variations and patterns, even if the sequences vary in speed or are out of phase. 

The DTW algorithm \cite{keogh2005exact} utilizes a dynamic programming approach to determine the optimal alignment between sequences, which minimizes the overall distance. This method results in a time complexity of $O(n^2)$, where $n$ represents the number of data points. For lengthy sequences and large databases, this complexity becomes computationally prohibitive and inefficient for retrieval purposes. To address this issue, we employ the FastDTW algorithm~\cite{salvador2004fastdtw}, which reduces the time complexity to approximately $O(n)$ by implementing a multi-level approach. The algorithm initiates by calculating DTW at a coarser resolution to establish a preliminary alignment. Subsequently, it refines this alignment at progressively finer resolutions, significantly decreasing the number of required calculations. The mathematical foundation of FastDTW is illustrated in the following recursive formula:
\begin{equation}
D_{\text{FastDTW}}(T_1, T_2) = D(L_1, L_2)
\end{equation}
\begin{equation}
D(i, j) \approx \min\left\{D(i-1, j-1), D(i-1, j), D(i, j-1) \right\} + d(i, j)
\end{equation}
Here, \(D_{\text{FastDTW}}(T_1, T_2)\) denotes the FastDTW distance between time series \(T_1\), with length \(L_1\), and time series \(T_2\), with length \(L_2\). \(D(i,j)\) represents the approximate cumulative distance between two sequences up to the \(i_{\text{th}}\) and \(j_{\text{th}}\) elements, respectively, while \(d(i, j)\) signifies the distance between the \(i_{\text{th}}\) element of the first sequence and the \(j_{\text{th}}\) element of the second sequence. This approximation approach maximizes the matching accuracy while significantly reducing the complexity, making it more suitable for time series retrieval in large databases.

\textbf{In-Context Learning.}
With the time series database constructed, \name retrieves the most similar $K_1$ normal time series and the $K_2$ anomalous time series from $\tilde{\mathcal{S}}$ and $\hat{\mathcal{S}}$ to a target time series data $T$ by leveraging the FastDTW algorithm for similarity computation:
\begin{align}
\tilde{\mathcal{K}} = \{\tilde{S}_i : i \in \underset{i}{\mathrm{argmin}_{K_1}} \{D_{\text{FastDTW}}(T, \tilde{S}_i)\}\} \\
\hat{\mathcal{K}} = \{\hat{S}_i : i \in \underset{i}{\mathrm{argmin}_{K_2}} \{D_{\text{FastDTW}}(T, \hat{S}_i)\}\}
\end{align}
Here, \(\tilde{\mathcal{K}}\) and \(\hat{\mathcal{K}}\) are the actual sets of \(K_1\) most similar normal and \(K_2\) most similar anomalous time series to \(T\), respectively. Incorporating the retrieved similar normal and abnormal data from historical records into the prompt of the LLMs enhances the ICL capability (see Figure~\ref{fig:prompt}). This provides a comprehensive representation of the data background, anomaly patterns, and distinctions tailored to a specific time series dataset. Consequently, the detection performance of the proposed method, denoted as \name, is improved without the necessity for fine-tuning, as demonstrated in many applications \cite{dong2022survey, zhang2024allhands, jiang2024xpert}.

\subsection{Anomaly Detection Chain of Thoughts \label{sec:prompt}}
\begin{figure*}[t]
\centering
\includegraphics[width=1\textwidth]{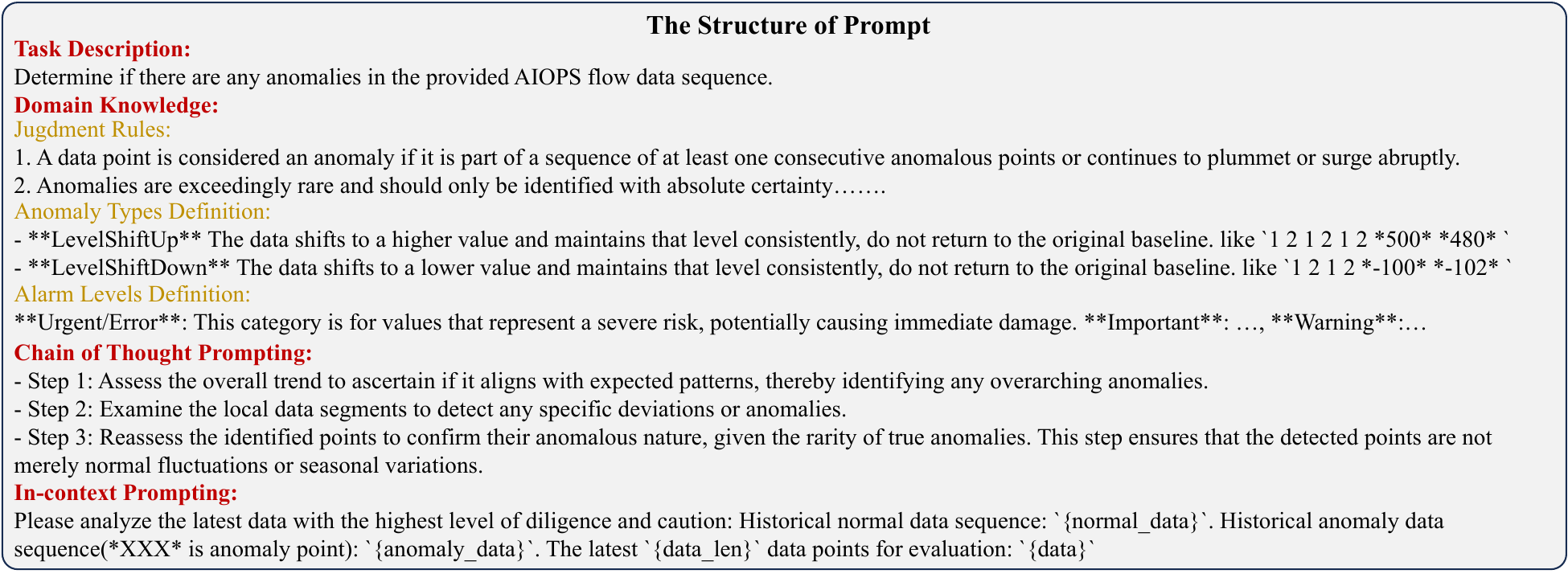}
\vspace*{-1em}
\caption{The overall structure of the prompt.}
\label{fig:prompt}
\end{figure*}

The CoT prompting technique guides LLMs to perform step-by-step reasoning, thereby enhancing the quality and consistency of the inference process \cite{wei2022chain}. We adapt this approach to the domain of TSAD by introducing the Anomaly Detection Chain of Thoughts (AnoCoT), which is specifically designed to address the context and logical thinking requirements for both anomaly detection and interpretation. This adaptation encompasses two key components: Domain Knowledge Injection and Expert Step-Wise Inference. We show the overall prompt structure that encloses AnoCoT in Figure~\ref{fig:prompt}.

\noindent\textbf{Domain Knowledge Injection} 
Domain knowledge comprises the specialized insights and expertise pertinent to a specific domain or dataset in the context of TSAD. This knowledge encompasses three main aspects: \emph{(i)} judgment rules, \emph{(ii)} anomaly type definitions, and \emph{(iii)} criteria for alarm levels. Judgment rules assist in establishing criteria that distinguish normal from abnormal patterns, tailored to a particular data background. This helps clarify the boundary between normal and abnormal data \cite{liu2023logprompt, gao2023chatgpt}. Anomaly type definitions and criteria for alarm levels serve as background knowledge of the anomalies, which not only improve the model's judgment but also contribute to the interpretability for subsequent use. The injected domain knowledge varies based on the context of the data, allowing users to tailor it to their specific needs and enhancing the flexibility of the approach.

By incorporating comprehensive domain knowledge, LLMs are better equipped to enhance their TSAD capabilities and adapt to specific data sets. This integration of expert knowledge allows the LLMs to generate more accurate and contextually relevant results, improving the overall performance and interpretability of the anomaly detection process.

\noindent\textbf{Expert Step-Wise Inference} 
With the domain knowledge, the AnoCoT method unfolds the inference in three distinct steps tailored to the TSAD. The first step involves a global trend assessment, which analyzes the overall trend within the time series, identifying potential indicators of gradual deterioration, such as level shifts. The second step conducts local anomaly assessment, is specifically designed to detect abrupt anomalies occurring on a smaller scale, such as spikes, which may signal sudden incidents. The third step takes reassessment, entails a re-evaluation of the anomalies identified in the preceding steps. Given the inherent infrequency of true anomalies and the high occurrence of false positives in extensive models, this step is indispensable. 

The step-wise inference design emulates the logical reasoning of expert engineers, which substantially enhances the accuracy of TSAD and the quality of interpretability. This approach allows the model to better mimic human decision-making processes, resulting in more reliable and easily understandable anomaly detection outcomes.

\subsection{Interpretable TSAD\label{sec:interpretation}}
Interpretability plays a crucial role in TSAD since it provides valuable information for decision-making. The proposed method \name, generates a comprehensive anomaly report in natural language to assist users in this regard. The report encompasses three key aspects: \emph{(i)} anomaly explanation, \emph{(ii)} anomaly type classification, and \emph{(iii)} alarm level classification.

\textbf{Anomaly Explanation}
In the anomaly explanation component, \name utilizes the AnoCoT to generate a step-wise explanation for its inference. This explanation synthesizes insights from an initial global trend assessment, which contextualizes anomalies within the broader data landscape, and detailed local anomaly assessments that identify specific irregularities, such as sudden spikes and re-evaluation results. This approach emulates the logical thinking flow of a human expert, which not only captures the most pertinent points for decision-making but also produces a human-readable report in natural language using LLMs. Such interpretability is not achievable with traditional TSAD methods.

\textbf{Anomaly Type Classification}
\begin{figure}[t]
\centering
\includegraphics[width=1\columnwidth]{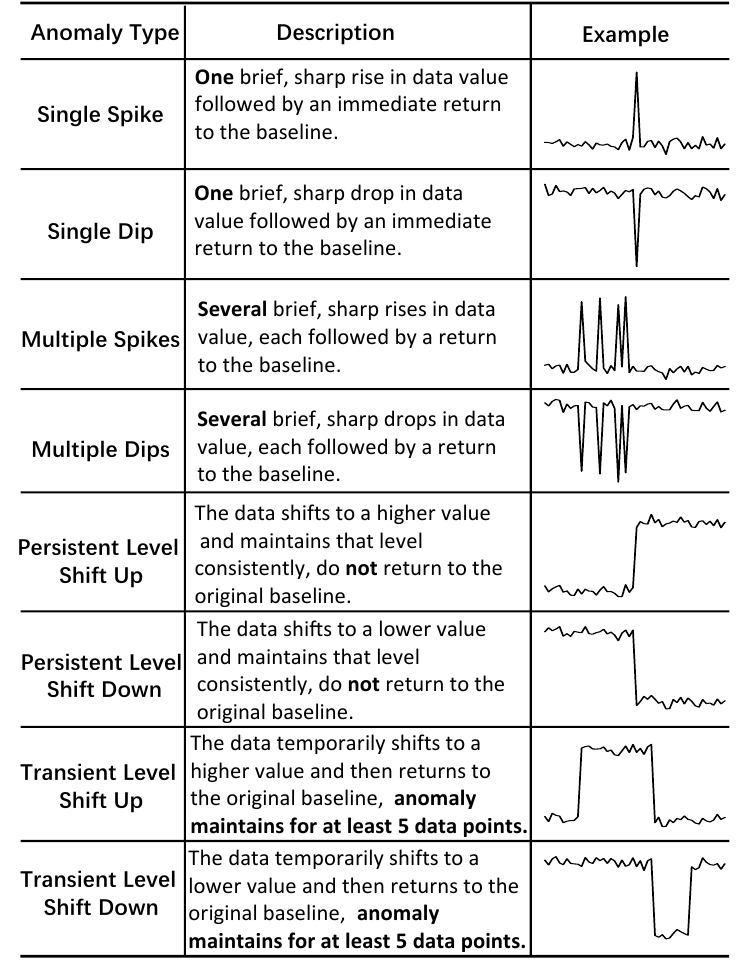}
\vspace*{-2em}
\caption{Selected anomaly types and definition employed in this study.}
\label{fig:anomaly}
\end{figure}
The identification of specific anomaly types can furnish engineers with additional information for pinpointing the root cause of issues \cite{blazquez2021review}. For example, abrupt anomalies, such as spikes, signify sudden incidents, while trend anomalies, like level shifts up, indicate deterioration. Although existing time series anomaly detection research focuses on identifying statistical anomalies, it often overlooks the classification of anomaly types.

In our study, we have recognized eight distinct and common anomaly types \cite{wu2021identifying}, as illustrated in Figure~\ref{fig:anomaly}, by conducting a pre-examination of the dataset. We incorporate their textual descriptions and examples within the prompt, as demonstrated in Figure\ref{fig:prompt}. These types include Single Spike, Single Dip, Multiple Spike, Multiple Dip, Transient Level Shift Up, Transient Level Shift Down, Persistent Level Shift Up, and Persistent Level Shift Down, which encompass the majority of patterns observed in the utilized dataset.

\textbf{Alarm Level Classification} 
Alarm levels represent the severity or urgency of detected anomalies, typically characterized by their potential impact on the system or process. Three distinct degrees of alarm levels have been established: Warning, Important and Urgent/Error. The ``Warning'' level signifies minor deviations that do not necessitate immediate action, while the ``Important'' level indicates conditions that could potentially escalate into future problems or cause system stress but are not immediately hazardous. In contrast, the ``Urgent/Error'' levels emphasize critical issues that demand immediate attention. These levels are defined in conjunction with the data context, and their prediction proves highly beneficial for subsequent decision-making and prioritization \cite{zhang2019deep}.

\section{Anomaly Detection Performance}
In this section, we first evaluate the TSAD performance of \name by examining the TSAD accuracy.

\subsection{Experiment Setup}
\begin{table}[t]
\centering
\caption{The statistics of time series anomaly detection dataset employed in the experiments.}
\label{tab:dataset}
\resizebox{\columnwidth}{!}{%
\begin{tabular}{@{}lrrrcc@{}}
\toprule
\textbf{Dataset} & \multicolumn{1}{r}{\textbf{All Points}} &
  \multicolumn{1}{r}{\textbf{Test Points}} &
  \multicolumn{1}{r}{\textbf{Anomalies}} &
  \textbf{Real$^*$} &
  \textbf{Synthetic$^*$} \\ \midrule
KPI   & 6,198,513 & 309,925 & 2.35\% &\ding{51}  &\ding{55}  \\
Yahoo & 738,866  & 369,433 & 0.71\%  &\ding{51}  &\ding{51}  \\
WSD   & 7,898,005 & 394,900 & 1.56\% &\ding{51}  &\ding{55}  \\ \bottomrule
\end{tabular}%
}
\end{table}

\textbf{Dataset.} 
We utilized three univariate time series datasets commonly employed in TSAD, to evaluate the performance. These include:
\begin{itemize}[leftmargin=*]
    \item \textbf{KPI} \cite{li2022constructing}: This dataset primarily consists of Key Performance Indicators (KPIs) from internet company business monitoring, such as website traffic and server CPU usage rates. These data are crucial for real-time monitoring of system health.
    \item \textbf{Yahoo} \cite{laptev2015benchmark}: Released by Yahoo Labs, this dataset includes both real-world and synthetic time series data suitable for anomaly detection and predictive model assessment. It encompasses a variety of anomaly types, such as point anomalies and contextual anomalies, providing a rich testing ground for algorithms.
    \item \textbf{WSD} \cite{zhang2022efficient}: This dataset focuses on anomaly detection in Web service performance indicators, such as response times and error rates. The data aids in monitoring the quality of Web services, quickly identifying issues like service delays and functional failures.
\end{itemize}
The details of each dataset are presented in Table~\ref{tab:dataset} \footnote{``Real\*'' indicates datasets derived from real-world data sources. ``Synthetic\*'' denotes datasets that have been artificially created to include specific anomalies.}. To assess the performance of TSAD, we selected the final 50\% of data from the Yahoo dataset and 5\% of data from both the KPI and WSD datasets as the test set. For the baseline models, we use the first 50\% of the data in each datasets for training and model selection.

\textbf{Baseline Models.} 
We compare our proposed \name against a diverse set of state-of-the-art TSAD methods, categorized by their detection mechanisms and typical applications. These include:
\begin{itemize}[leftmargin=*]
    \item \textbf{SPOT} \cite{siffer2017anomaly}: A statistical method grounded in extreme value theory, ideal for outlier detection in univariate data streams.
    \item \textbf{SRCNN} \cite{ren2019time}: A supervised learning approach that uses convolutional neural networks to leverage spatial dependencies.
    \item \textbf{DONUT} \cite{xu2018unsupervised}: An unsupervised anomaly method utilizing variational autoencoder to denoise the anomalies and further learn the robust representation of normal patterns.
    \item\textbf{VQRAE} \cite{kieu2022anomaly}: An unsupervised anomaly method that uses detection variational quasi-recurrent autoencoders.
    \item \textbf{AnoTransfer} \cite{zhang2022efficient}: An approach that enhances the VAE framework with transfer learning to adapt quickly to different settings in an unsupervised manner.
    \item \textbf{Informer} \cite{zhou2021informer}: A predictive model that utilizes advanced attention mechanisms and transformer architectures to forecast and detect deviations from normal sequences.
    \item \textbf{TFAD} \cite{zhang2022tfad}: A method that integrates traditional feature engineering with anomaly detection strategies in a supervised learning framework.
    \item\textbf{Anomaly-Transformer} \cite{xu2021anomaly}: An unsupervised method that leverages the transformer architecture with an anomaly-attention mechanism to compute the association discrepancy.
    \item\textbf{LLMTime} \cite{gruver2024large}: A zero-shot LLM-based method to forecast time series, and we use the prediction error as anomaly signals.    
\end{itemize}
\fix{There are baseline missing. And the order should match table 2.}
This comprehensive comparison spans from traditional statistical approaches to cutting-edge deep learning models, ensuring a thorough analysis of their efficacy in anomaly detection within time-series data. We evaluated the performance of the baseline models by adhering to their original code repository and configuration settings. In the case of our \name, we chose GPT-4-1106-preview as the primary language modeling model.


\begin{table*}[t]
\centering
\caption{The overall experimental results across three TSAD datasets. \textbf{Bold} numbers denote the best performance among all models, \underline{underlined} signify the second-best performance.}
\label{tab:main}
\begin{tabular}{lcccccccc}
\hline
 & \multicolumn{2}{c}{\textbf{KPI}}  & \multicolumn{2}{c}{\textbf{WSD}}  & \multicolumn{2}{c}{\textbf{Yahoo}} & \multicolumn{2}{c}{\textbf{Average}} \\ 
\cline{2-9} 
\multirow{-2}{*}{\textbf{Method}} & \textbf{Best F1} & \textbf{Delayed F1} & \textbf{Best F1} & \textbf{Delayed F1} & \textbf{Best F1} & \textbf{Delayed F1} & \textbf{Best F1} & \textbf{Delayed F1} \\ 
\hline

SPOT               & 0.269 & 0.269 & 0.273 & 0.273 & 0.417 & 0.417 & 0.320 & 0.320 \\
SRCNN              & 0.617 & 0.488 & 0.286 & 0.231 & 0.251 & 0.198 & 0.385 & 0.306 \\
DONUT              & 0.382 & 0.294 & 0.224 & 0.141 & 0.215 & 0.215 & 0.274 & 0.217 \\
VQRAE              & 0.272 & 0.137 & 0.179 & 0.189 & 0.510 & 0.492 & 0.320 & 0.273 \\
Anotransfer        & 0.685 & 0.461 & \underline{0.674} & 0.379 & 0.567 & 0.496 & 0.642 & 0.445 \\
Informer           & \underline{0.859} & \textbf{0.715} & 0.534 & 0.370 & 0.707 & 0.671 & 0.700 & 0.585 \\
TFAD               & 0.751 & 0.631 & 0.644 & \underline{0.396} & \textbf{0.779} & \textbf{0.775} & \underline{0.725} & \textbf{0.601} \\
Anomaly Transformer & \textbf{0.918} & 0.336 & 0.670 & 0.092 & 0.274 & 0.029 & 0.621 & 0.152 \\
LLMTIME            & 0.333 & 0.194 & 0.029 & 0.022 & 0.023 & 0.023 & 0.128 & 0.080 \\
\cmidrule{1-9}
\name     & 0.843 & \underline{0.667} & \textbf{0.711} & \textbf{0.401} & \underline{0.724} & \underline{0.695} & \textbf{0.759} & \underline{0.588} \\ 
\hline
\end{tabular}
\end{table*}

\begin{figure}[t]
\centering
\includegraphics[width=0.8\columnwidth]{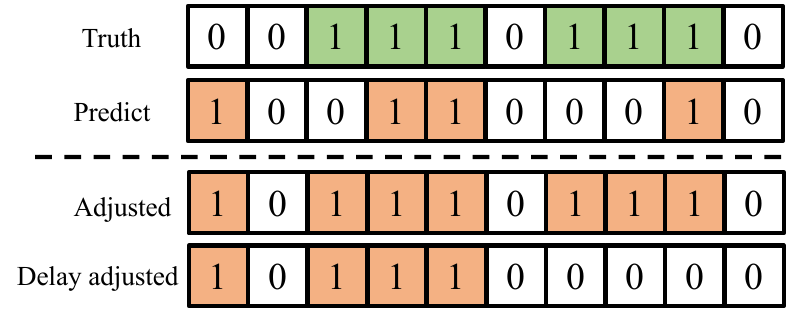}
\caption{An illustration of the delay F1.}
\label{fig:delay}
\end{figure}
\textbf{Evaluation Metrics.} In practical applications, operators prioritize the detection of continuous anomalous segments in time-series data over point-wise anomaly detection, which involves classifying each individual data point as an anomaly. This shift in focus is attributed to the significant impact that anomalous segments can potentially have. To address these specific needs, our research employs two metrics: Best F1 and Delayed F1, drawing inspiration from DONUT \cite{xu2018unsupervised} and SRCNN \cite{ren2019time}, respectively.


The \textbf{Best F1} is designed to measure the peak F1 score that a model can achieve across different thresholds, incorporating a point adjustment strategy. This provides a benchmark for the optimal performance capability of the model under ideal conditions. In \name, we adjust the threshold by considering the number of detected anomaly points as a variable. \textbf{Delayed F1} score caters to the practical necessity of prompt anomaly detection by introducing a delay threshold, denoted as $K$. If an anomaly segment is identified after surpassing this threshold $K$, it is deemed as undetected. This metric modifies predictions by considering a specific delay. The example shown in Figure~\ref{fig:delay}, where the delay $K = 1$,  the second anomaly segment is prolonged to encompass three points due to the delay adjustment, whereas the third segment is overlooked because its detection calls for two-time intervals, exceeding the maximum delay threshold of 1. This approach balances the importance of swift detection against the imperative of maintaining detection accuracy, a balance that is especially crucial in scenarios where prompt detection can significantly reduce potential risks or losses. For the WSD and KPI, the delay is configured at 7, and for Yahoo datasets, it is set to 3.


\subsection{Performance Comparison}
In Table~\ref{tab:main}, we report the results for Best F1 and Delayed F1 on different datasets and methods. Observe that \name on average achieves the best performance on Best F1 and second-best performance on Delayed F1 across the three datasets, underscoring its superior performance in accuracy and timeliness. These results suggest that the LLM based approach can deliver reliable TSAD compared to traditional methods, opening a new methodology for the TSAD domain. 

Examining the breakdown performance, \name ranks in the top three for both F1 and Delayed F1 for all datasets, demonstrating its robust performance. The performance of various baseline methods on datasets exhibits variance. For instance, SPOT generally underperforms across most datasets due to its erroneous treatment of outliers as anomalies, which do not always manifest in such a manner. Anomaly Transformer achieves commendable results in terms of the highest Best F1 on most datasets but demonstrates a low Delayed F1. It detects anomalies based on their relation to nearby points, effectively capturing correlations only when the anomalous points are relatively central within a window. On the other hand, LLMTIME, which also employs LLM for prediction and uses the error as an anomaly indicator, has very low performance. This observation suggests the superior pipeline of \name. By introducing ICL and AnoCoT, \name utilizes LLM more effectively, making it more suitable for the TSAD task.

It has been observed that \name does not always achieve the best results in certain datasets. For instance, in the KPI dataset, this is due to the presence of long-duration anomalies that may extend across multiple windows. As a result, \name may struggle in these cases due to the limitation of window size, while other methods are trained on the entire dataset and have incorporated such long-term knowledge. In the case of the Yahoo dataset, TFAD performs better. We have noticed that the dataset annotations sometimes only mark critical anomaly points, such as the point of a level shift, without indicating subsequent sustained anomalies. Consequently, \name may identify the sustained values post-shift as anomalies, leading to slightly lower precision. Nevertheless, \name still ranks among the top three baselines in these two datasets and the best on average, making it a reliable method.




\begin{table}[h]
\centering
\caption{Performance comparison of different settings of ICL.}
\label{tab:icl}
\resizebox{\columnwidth}{!}{%
\begin{tabular}{@{}cccccccc@{}}
\toprule
\multirow{2}{*}{\textbf{Sample}} &
  \multirow{2}{*}{\textbf{Number}} &
  \multicolumn{2}{c}{\textbf{WSD}} &
  \multicolumn{2}{c}{\textbf{KPI}} &
  \multicolumn{2}{c}{\textbf{Yahoo}} \\ \cmidrule(l){3-8} 
 &
   &
  \textbf{Best F1} &
  \textbf{Delayed F1} &
  \textbf{Best F1} &
  \textbf{Delayed F1} &
  \textbf{Best F1} &
  \textbf{Delayed F1} \\ \midrule
\multirow{5}{*}{Dynamic} & 1 pos         & 0.548          & 0.304          & 0.756          & 0.305          & 0.602          & 0.602          \\
                         & 1 neg         & 0.595          & 0.327          & 0.768          & 0.431          & 0.622          & 0.622          \\
                         & 1 pos/1 neg   & 0.653          & 0.335          & 0.816          & 0.641          & 0.679          & 0.655          \\
                         & 2 pos/1 neg   & 0.711          & 0.401          & \textbf{0.843} & 0.667          & \textbf{0.724} & \textbf{0.695} \\
                         & 4 pos/1 neg   & \textbf{0.716} & \textbf{0.409} & 0.812          & \textbf{0.697} & 0.691          & 0.669          \\ \midrule
Fixed                    & 4 pos/1 neg   & 0.547          & 0.351          & 0.733          & 0.283          & 0.638          & 0.602          \\ \midrule
Zero-shot                & -             & 0.512          & 0.270          & 0.711          & 0.277          & 0.574          & 0.563          \\ \bottomrule
\end{tabular}%
}
\end{table}

\subsection{Ablation Analysis}
To delve into the effect of individual components in \name, we explore the impact of three distinct factors: \emph{(i)} in-context learning, \emph{(ii)} chain-of-thoughts prompting,  \emph{(iii)} the injection of domain knowledge, and \emph{(iv)} the foundation LLM inference engine.

\subsubsection{In-Context Learning}
First, we evaluate the effectiveness of ICL in \name along two dimensions: \emph{(i)} whether to retrieve samples dynamically, use fixed ones, or not use ICL at all (zero-shot \cite{kojima2022large}), and \emph{(ii)} the number of positive (anomaly) and negative (normal) samples used for ICL.

Table~\ref{tab:icl} presents the performance comparison of different ICL settings in \name using GPT-4. Observe that using few-shot ICL can significantly boost the performance of TSAD, regardless of whether the samples are fixed or dynamic. This demonstrates the superiority of ICL. By providing demonstrations via examples, \name establishes a clearer boundary between normal and abnormal patterns, boosting its performance. On the other hand, it appears that using samples retrieved dynamically achieves better performance than fixed ones. This suggests that using similar samples can better help LLMs distinguish ambiguous cases.

Taking a closer look at the effect of the number of samples retrieved, it appears that the performance of \name increases with the number of samples, but this increase becomes flattened if more than 2 positive/1 negative samples are provided. This observation makes sense, as more samples can, to some extent, provide rich patterns and information for \name, but continually increasing the number of samples may lead to repetitive information. Using the setting of 2 positive/1 negative samples can be a good trade-off between complexity and performance.



\begin{figure}[t]
\centering
\includegraphics[width=1\columnwidth]{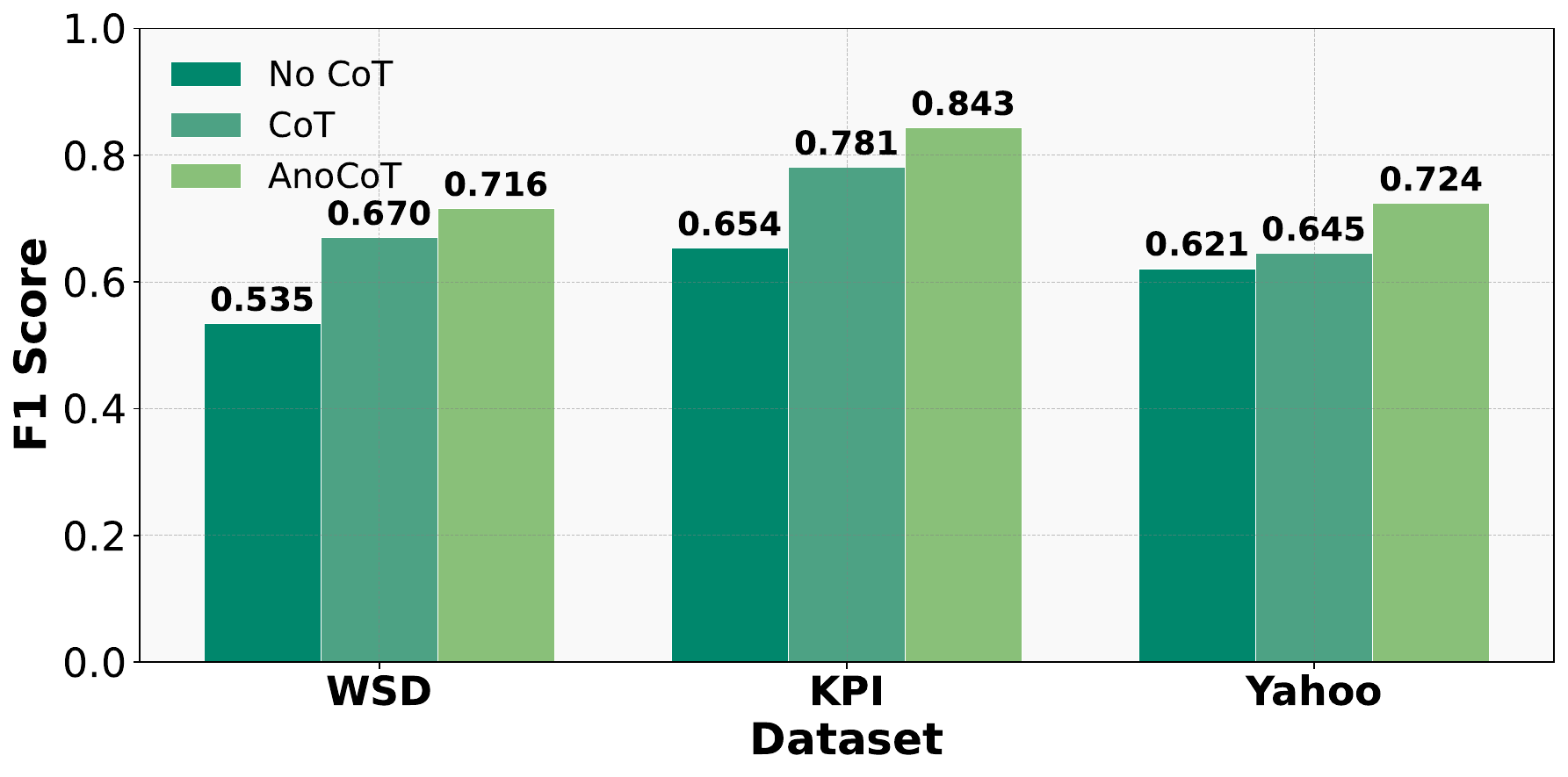}
\caption{Performance comparison of different prompting schemes.}
\label{fig:cot}
\end{figure}

\begin{table}[t]
\centering
\caption{Comparison of \name on using different LLMs.}
\label{tab:model_performance}
\resizebox{\columnwidth}{!}{%
\begin{tabular}{lcccccc}
\hline
\multirow{2}{*}{\textbf{Model}} & \multicolumn{2}{c}{\textbf{WSD}} & \multicolumn{2}{c}{\textbf{KPI}} & \multicolumn{2}{c}{\textbf{Yahoo}} \\ \cline{2-7} 
                      & \textbf{Best F1} & \textbf{Delayed F1} & \textbf{Best F1} & \textbf{Delayed F1} & \textbf{Best F1} & \textbf{Delayed F1} \\ \hline
\textbf{Llama-3-70B}  & 0.360 & 0.264             & 0.701             & 0.289          & 0.490             & 0.471             \\
\textbf{GPT-3.5}               & 0.233 & 0.069 & 0.409             & 0.134         & 0.204             & 0.194             \\
\textbf{GPT-4}       & \textbf{0.716}    & \textbf{0.409}   & \textbf{0.843}    & \textbf{0.667} & \textbf{0.724}    & \textbf{0.695}    \\
 \hline
\end{tabular}%
}
\end{table}
\subsubsection{Chain-of-Thoughts Prompting.} 
Now we move to evaluate the effectiveness of our AnoCoT by comparing its performance with \name without CoT and standard CoT (asking LLM to think step-by-step), as shown in Figure~\ref{fig:cot}. As expected, CoT can significantly improve the performance of TSAD in terms of Best F1 on all datasets, while our AnoCoT, which incorporates domain knowledge and follows the logical thinking of experts, can further boost the performance to a greater level. On average, CoT improves the Best F1 by 9.5\%, and using AnoCoT can further improve CoT by 6.2\%, which is a substantial boost. This improvement can be attributed to the fact that as LLMs infer like humans, using step-wise thinking combined with injecting domain expertise leads to improved performance in TSAD.

\subsubsection{Domain Knowledge Injection.} 
We also delve into the effectiveness of domain knowledge to see how its injection can help with TSAD. Table~\ref{tab:ablation} shows the impact of removing one of the domain knowledge components defined in Section~\ref{sec:prompt}. Observe that removing any domain knowledge component will lead to a performance drop, as expected. The drop is most significant in the absence of specific judgment rules. This underscores the importance of defining textual rules for LLM's understanding of the definition of anomalies. Although type and alarm level do not directly provide guidance for TSAD, they are also important, as they act as side information to deliver the context of the task. This enhances the ability of \name, making it more sophisticated in the task.


\subsubsection{Foundation LLM\label{sec:llm}}
Finally, we evaluated the performance of \name using different LLM engines to compare their divergence. Table \ref{tab:model_performance} shows the results for GPT-3.5 \cite{openai2020gpt3}, GPT-4 \cite{achiam2023gpt}, and Llama-3-70B-Instruct \cite{meta2024llama3}. It can be observed that GPT-4 outperforms the other two models significantly in all datasets, while Llama-3-70B-Instruct demonstrates superior performance compared to GPT-3.5. Considering that the TSAD task is heavily dependent on domain-specific knowledge and the ability to follow instructions and comprehend content, only GPT-4 fulfills the necessary criteria. This underscores the vital role of advanced LLM models in accomplishing such tasks.


\begin{table}[]
\centering
\caption{Ablation studies for domain knowledge injection.}
\label{tab:ablation}
\resizebox{\columnwidth}{!}{%
\begin{tabular}{lcccccc}
\hline
\textbf{Method} & \multicolumn{2}{c}{\textbf{WSD}} & \multicolumn{2}{c}{\textbf{KPI}} & \multicolumn{2}{c}{\textbf{Yahoo}} \\ \cline{2-7} 
                & \textbf{Best F1} & \textbf{Delayed F1} & \textbf{Best F1} & \textbf{Delayed F1} & \textbf{Best F1} & \textbf{Delayed F1} \\ \hline
\name           & \textbf{0.716}   & \textbf{0.409}      & \textbf{0.843}   & \textbf{0.667}      & \textbf{0.724}   & \textbf{0.695}      \\
\quad\textbf{-w/o rule}  & 0.423            & 0.228             & 0.652            & 0.383             & 0.555            & 0.550             \\
\quad\textbf{-w/o type}  & 0.565            & 0.378             & 0.775            & 0.449             & 0.643            & 0.631             \\
\quad\textbf{-w/o level} & 0.696            & 0.353             & 0.810            & 0.628             & 0.700            & 0.688             \\
\hline
\end{tabular}%
}
\end{table}

\section{Interpretability Evaluation}
We proceed to analyze the quality of anomaly interpretations generated by \name from three distinct perspectives: anomaly explanation, anomaly type classification, and the incorporation of urgency level into the anomaly explanation for joint analysis.

\subsection{Experiment Setup}
The original dataset is solely labeled with anomaly points and lacks ground truth information for interpretability evaluation. To address this limitation, we employ \name to generate output for each type of interpretation and enlist the expertise of 5 DevOps engineers to manually assess the quality of interpretation in each case. With an average of 3 years of TSAD experience, these engineers offer a well-rounded perspective on the evaluation process.

\subsubsection{Explanation Evaluation Setup}
We combine the evaluation of anomaly explanation and urgency level, as assessing the urgency level independently, without the context and accompanying interpretation, proves to be challenging. To thoroughly assess these aspects of \name's performance, we adopt two primary criteria: \textbf{usefulness} and \textbf{readability} \cite{liu2023logprompt}. Usefulness refers to the effectiveness and relevance of an explanation in practical applications, with a score range between 1 and 5. Readability focuses on the clarity and ease of understanding of the explanation, scored between 1 and 3. More details on each score can be found in Appendix \ref{interpret_def}. These criteria are crucial for determining how effectively the explanations generated by \name contribute to an improved understanding of detected anomalies and provide actionable insights.

We randomly selected 100 anomaly cases in total from the three datasets for human evaluation and compared the performance of the GPT-4 version of \name using standard CoT and AnoCoT methods. In addition to the mean and standard deviation (std) scores from each rater, we also provide the evaluation of High Incidence Proportion (HIP) \cite{liu2023logprompt}, which calculates the percentage of evaluations surpassing a satisfying performance threshold. The threshold is set at a score of 4 for usefulness and 2 for readability.



\subsubsection{Anomaly Type Classification Setup} 
We also randomly selected 100 anomalous samples in total to manually evaluate the performance of anomaly type classification. Each sample was asked to be labeled with one or multiple types of anomalies. To evaluate the classification performance, we employ two metrics: Acc (any-hit)~\cite{zhou2012multi} and Micro F1 \cite{hastie2009elements}.
\textbf{Acc (any-hit)} calculates the ratio of correctly predicted anomaly types to the total number of samples. A type prediction is considered a ``hit'' if it matches any human label. \textbf{Micro F1} is a metric that aggregates the contributions of all classes to compute the average metric. This makes it robust to class imbalance and emphasizes the importance of correctly classifying each instance. \fix{more detail}. By using these metrics, we can effectively assess the performance of \name in classifying anomaly types.

\begin{table}[]
\centering
\caption{Results of human evaluation for explanation generated by \name.}
\resizebox{\columnwidth}{!}{%
\begin{tabular}{@{}ccccccccccccc@{}}
\toprule
\multirow{3}{*}{\textbf{Raters}} & \multicolumn{6}{c}{\textbf{AnoCoT}}            & \multicolumn{6}{c}{\textbf{Standard CoT}}      \\ \cmidrule(l){2-13} 
 &
  \multicolumn{3}{c}{\textbf{Usefulness}} &
  \multicolumn{3}{c}{\textbf{Readability}} &
  \multicolumn{3}{c}{\textbf{Usefulness}} &
  \multicolumn{3}{c}{\textbf{Readability}} \\ \cmidrule(l){2-13} 
 &
  \textbf{Mean} &
  \textbf{Std} &
  \textbf{HIP} &
  \textbf{Mean} &
  \textbf{Std} &
  \textbf{HIP} &
  \textbf{Mean} &
  \textbf{Std} &
  \textbf{HIP} &
  \textbf{Mean} &
  \textbf{Std} &
  \textbf{HIP} \\ \midrule
\textbf{Rater1}                  & 4.45 & 0.95 & 78.35\% & 2.98 & 0.14 & 100.00\% & 3.85 & 1.56 & 62.89\% & 2.93 & 0.30 & 98.97\%  \\
\textbf{Rater2}                  & 4.22 & 0.93 & 76.29\% & 2.37 & 0.63 & 91.75\%  & 3.79 & 1.03 & 63.92\% & 2.05 & 0.55 & 87.63\%  \\
\textbf{Rater3}                  & 4.09 & 0.98 & 79.38\% & 2.94 & 0.24 & 100.00\% & 3.57 & 0.86 & 56.70\% & 2.94 & 0.24 & 100.00\% \\
\textbf{Rater4}                  & 3.80 & 1.04 & 64.95\% & 2.31 & 0.47 & 98.97\%  & 3.29 & 1.43 & 49.48\% & 2.30 & 0.66 & 88.66\%  \\
\textbf{Rater5}                  & 3.75 & 1.23 & 62.89\% & 2.70 & 0.62 & 94.85\%  & 3.41 & 1.19 & 47.42\% & 2.65 & 0.52 & 97.81\%  \\
\cmidrule{2-13}
\textbf{Average}                 & 4.06 & 1.03 & 72.37\% & 2.66 & 0.42 & 97.11\%  & 3.58 & 1.21 & 56.08\% & 2.57 & 0.45 & 94.61\%  \\ \bottomrule
\end{tabular}%
}
\label{tab:explanation}
\end{table}

\subsection{Evaluation Results}

\subsubsection{Explanation Evaluation}
Table~\ref{tab:explanation} presents a comparison of AnoCoT and standard CoT based on human evaluation of anomaly explanations generated by \name. It can be observed that AnoCoT outperforms standard CoT in terms of both usefulness and readability across all raters, with an average increase of 13.4\% in usefulness and 3.5\% in readability. This result suggests that AnoCoT, which incorporates domain knowledge for TSAD and follows expert reasoning flow, consistently provides better quality explanations for anomaly prediction, delivering more insightful, detailed, useful, and user-friendly information for engineers in decision-making processes.

Furthermore, AnoCoT achieves 72.37\% and 97.11\% HIP for usefulness and readability, respectively, indicating that \name can deliver satisfactory explanations for a majority of samples, making it practical for real industry scenarios. Overall, these findings suggest that AnoCoT not only enhances the performance of TSAD but also improves the transparency and interpretability of the anomaly detection process, making it a highly valuable addition to the task.

\subsubsection{Anomaly Classification Evaluation}
In Table~\ref{tab:type_classification}, we present the performance comparison of anomaly classification across three datasets for different LLM models\footnote{``Acc'' indicates the Acc (any-hit) metric}. It can be observed that GPT-4 consistently outperforms other baselines as the most reliable model for anomaly classification across all datasets. It achieves over 90\% accuracy on WSD and Yahoo datasets, and 79\% on the KPI dataset, while also obtaining high Micro F1 scores for all datasets. This performance is remarkable, given that the samples are not labeled with anomaly type labels. GPT-4 can accurately classify anomalies based solely on the textual description of types, which demonstrates its remarkable interaction-following ability. This high performance suggests that \name can deliver excellent anomaly classification quality to assist interpretable TSAD and decision-making processes. However, this advantage is limited to GPT-4. The task poses high requirements on the LLM model, as confirmed in Section~\ref{sec:llm}.

\begin{table}[]
\centering
\caption{Results of the evaluation for anomaly type classification.}
\resizebox{\columnwidth}{!}{%
\begin{tabular}{@{}lllllll@{}}
\toprule
\multicolumn{1}{c}{\multirow{2}{*}{\textbf{Model}}} &
  \multicolumn{2}{c}{\textbf{WSD}} &
  \multicolumn{2}{c}{\textbf{KPI}} &
  \multicolumn{2}{c}{\textbf{Yahoo}} \\ \cmidrule(l){2-7} 
\multicolumn{1}{c}{} &
  \textbf{Acc$^*$} &
  \textbf{Micro F1} &
  \textbf{Acc$^*$} &
  \textbf{Micro F1} &
  \textbf{Acc$^*$} &
  \textbf{Micro F1} \\ \midrule
\textbf{Llama-3-70B} &
   \multicolumn{1}{c}{0.379 } &
  \multicolumn{1}{c}{0.400 } &
  \multicolumn{1}{c}{0.286 } &
  \multicolumn{1}{c}{0.250 } &
  \multicolumn{1}{c}{0.600 } &
  \multicolumn{1}{c}{0.552 } \\

\textbf{GPT-3.5} &
   \multicolumn{1}{c}{0.310 } &
  \multicolumn{1}{c}{0.321 } &
  \multicolumn{1}{c}{0.143 } &
  \multicolumn{1}{c}{0.163 } &
  \multicolumn{1}{c}{0.375 } &
  \multicolumn{1}{c}{0.380 } \\
\textbf{GPT-4} &
  \multicolumn{1}{c}{0.900} &
  \multicolumn{1}{c}{0.776} &
  \multicolumn{1}{c}{0.790} &
  \multicolumn{1}{c}{0.647} &
  \multicolumn{1}{c}{0.930} &
  \multicolumn{1}{c}{0.804} \\
   \bottomrule
\end{tabular}%
}
\label{tab:type_classification}
\end{table}

\subsection{Case Study}
\begin{figure*}[htbp]
\centering
\includegraphics[width=1\textwidth]{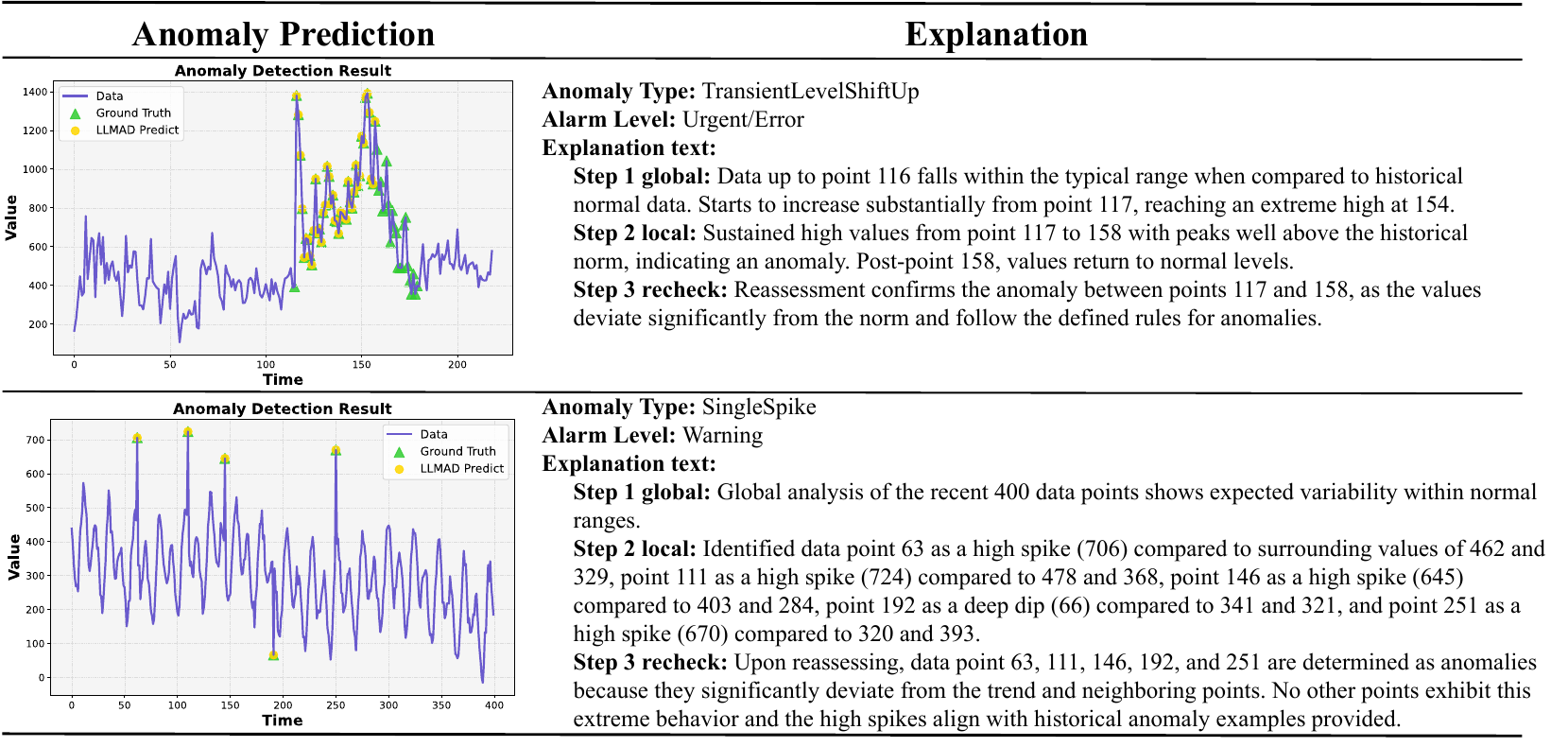}
\caption{Two case study of anomaly interpretation delivered by \name.}
\label{fig:cases}
\end{figure*}

We present two case studies in Figure~\ref{fig:cases} to demonstrate how \name successfully identifies and delivers interpretable TSAD results.

\textbf{Case 1: Transient Level Shift Detection}
In the first case study, we observe that the data exhibits a sudden and significant increase in data values, starting at point 117 and peaking at point 154. \name successfully identifies this pattern as a "Transient Level Shift Up" anomaly and captures the event range. This classification was enhanced by a comprehensive analysis that included:
\emph{(i)} A global assessment confirming that the shift was well above historical norms.
\emph{(ii)} A local trend analysis showing deviation from normal behaviors.
\emph{(iii)} A recheck validating the anomaly by demonstrating that the peak values deviated substantially from typical data trends.
Such interpretations are helpful and clear for engineers to localize the event and take further actions, highlighting the practical value of \name in providing interpretable TSAD results.

\textbf{Case 2: Singular Data Spike Detection}
The second case study presents a time series with regular fluctuations, where some peak values exceed normal ranges. \name successfully detects these spikes, which could indicate critical transient events. In its explanation, \name Pinpoints the specific anomalous points, Delivers reasoning steps by comparing them with normal patterns, and Generates a useful and human-readable report.
This enhances the system's usability for operators without specialized data science knowledge. Since the deviation is not significant, \name assigns this case a ``Warning'' level. This classification not only identifies issues but also aids in prioritizing system responses, ensuring that resources are allocated effectively and that operators can act on the most critical anomalies first. 

Overall, these two practical demonstrations show \name's value in providing interpretable TSAD results and facilitating informed decision-making.

\section{Cost and Latency Analysis\label{sec:cost}}
\begin{table}[t]
\centering
\caption{Cost and latency analysis for \name. \label{tab:cost}}
\label{tab:cost}
\begin{adjustbox}{width=0.34\textwidth,center}
\begin{tabular}{@{}lrrrr@{}}
\toprule
\multicolumn{1}{c}{\multirow{2}{*}{\textbf{Cost Items}}} & \multicolumn{2}{c}{\textbf{AnoCoT}} & \multicolumn{2}{c}{\textbf{CoT}} \\ \cmidrule(l){2-5} 
\multicolumn{1}{c}{}                                    & \textbf{Mean}       & \textbf{Std}       & \textbf{Mean}       & \textbf{Std}      \\ \midrule
Input Tokens                                            & 4,208 & 646          & 4,115 & 652 \\
Output Tokens                                           & 281  & 122          & 189  & 90  \\ \hline
API Calling Time (s)                                    & 14.91 & 7.59        & 10.52 & 5.07 \\
Total Time (s)                                          & 17.03 & 7.71        & 13.34 & 11.26 \\ \hline
Daily Cost (USD)                                        & 0.18  & 0.04        & 0.17  & 0.03 \\ \bottomrule
\end{tabular}
\end{adjustbox}
\end{table}

It is well-known that LLMs have a considerable number of parameters, which may result in high inference latency. In this section, we evaluate these aspects to provide essential information regarding the overhead of employing the proposed method, \name. We utilize the GPT-4-turbo-1106 model as the base and its variants under two methodologies: standard CoT and AnoCoT. Both methodologies were tested with a window length of 400, collecting data at a rate of one point per minute, consistent with the \textbf{highest sampling rate} in the WSD dataset among the three. In contrast, the sampling rates for the Yahoo and KPI datasets are one hour and 1 to 5 minutes, respectively.

We present the evaluation results in Table~\ref{tab:cost}. The mean and standard deviation (Std) values are computed across all samples from three datasets. We first compare the token usage between AnoCoT and CoT. It can be observed that AnoCoT introduces a minor increase in input and output tokens ($<100$) compared to CoT. A higher output token count implies a more comprehensive interpretation. Considering the improvements in both TSAD performance and interpretability, this additional token usage is justifiable. The proposed AnoCoT method requires, on average, 14.91 seconds for calling the GPT-4 API and 17.03 seconds for the total time per request. Given that each request handles 400 data points, corresponding to over a 6-hour time window, such latency is negligible.

\textbf{Financial Cost Estimation:} Daily costs were calculated based on the number of API calls needed to sample one data point per minute over 24 hours. In the non-overlapping setting, we computed the average token usage per day and determined the cost using the price of GPT-4-turbo-1106 \footnote{OpenAI. (2024). Pricing. Retrieved from \texttt{https://openai.com/pricing}. Specifically, the cost is \$0.01 per 1,000 tokens for input and \$0.03 per 1,000 tokens for output.}. The daily operational costs were approximately \$0.18 for AnoCoT and \$0.17 for standard CoT. Extrapolating these figures to an annual perspective, the cost amounts to approximately \$65.70 for AnoCoT and 
\$62.05 for standard CoT annually. AnoCoT offers more detailed outputs, it does so at the cost of higher token usage and longer processing times. Despite these variances, the annual costs for both methodologies remain moderate, reinforcing the viability of \name for long-term deployment in TSAD tasks, particularly when interpretability is a crucial requirement.

\section{Related work}
In this section, we review relevant research practices in TSAD and the use of LLMs for time series analysis.

\subsection{Time Series Anomaly Detection}

Time series anomaly detection utilizes a broad range of methodologies. Among these, SPOT~\cite{siffer2017anomaly} applies extreme value theory to detect anomalies in data streams by examining the statistical tails where extreme values are more likely to occur. As a representative supervised learning approach, SRCNN~\cite{ren2019time} combines spectral residuals with a CNN architecture to create an effective classifier for TSAD. TFAD~\cite{zhang2022tfad} incorporates time-frequency analysis within a decomposition framework to improve detection capabilities. Unsupervised methods are also popular in this area. For example, Donut~\cite{xu2018unsupervised} uses a variational autoencoder to detect anomalies in seasonal KPIs. Imdiffusion~\cite{chen2023imdiffusion} leverages diffusion models to reconstruct time series and uses imputation error as the anomaly identifier.

While these approaches may perform well in specific scenarios, they usually require massive amounts of data for training and lack interpretability, which is particularly important in the TSAD domain.

\subsection{Large Language Models for Time Series}
LLMs have increasingly demonstrated their potential in the field of time series analysis~\cite{su2024large,jin2023large}. Time-LLM~\cite{jin2023time} has activated LLM capabilities in time series through a novel embedding method that aligns data tokenization and encoding, as well as creating prompts to guide LLMs in time series analysis. LLMTime~\cite{gruver2024large} has shown that LLMs can effectively conduct zero-shot time series forecasting with appropriate pre-processing. Several research studies leverage LLMs to provide interpretable time series analysis. For example, the study~\cite{yu2023temporal} experiments with zero-shot/few-shot inference using GPT-4 and instruction-based fine-tuning with Llama~\cite{touvron2023llama} to generate explainable forecasts. Research in \cite{jin2024position} suggests using LLMs to provide accountability and transparency for time series analysis.

Although this area is becoming more robust, using LLMs for TSAD remains unexplored. We are pioneers in this direction, aiming to deliver accurate and interpretable TSAD using LLMs.

\section{Limitations}
Although \name excels in both accuracy and interpretability, it exhibits several limitations. In general, \name has higher inference latency and cost compared to traditional methods due to its reliance on LLMs. In scenarios that require real-time inference, \name may not be the most suitable choice. However, considering the cost analysis provided in Section~\ref{sec:cost}, the latency and cost are highly acceptable if the data sample size is not too large. Furthermore, \name offers the remarkable additional benefit of interpretability, which cannot be provided by traditional methods

In addition, while our experiments demonstrate the effectiveness of \name across multiple TSAD datasets, they rely on well-crafted prompts. Creating these prompts benefits from accurate domain knowledge, which ideally should be provided by experienced engineers. This knowledge is necessary to ensure good TSAD performance and interpretability. However, this requirement is a one-time effort, as \name can automatically generate output anomalies and explanations without further human involvement once the domain knowledge has been incorporated. This characteristic makes \name a practical and valuable method for a wide range of applications, balancing the need for expert input with the advantages of automated analysis and interpretation.

\section{Conclusion}
In this paper, we introduced \name, a novel approach leveraging LLMs to deliver accurate and interpretable time series anomaly detection with low cost. \name retrieves similar samples from historical data to enable ICL to contextualize the dataset and the task of TSAD. It also uses AnoCoT to inject domain knowledge tailored to a specific dataset and instructs \name to think like a human expert for TSAD. These techniques significantly improve the performance and quality of interpretability of TSAD, making the task more effective. Our experimental results on public datasets demonstrate that \name not only achieves competitive performance compared to state-of-the-art methods in terms of Best F1 but also excels in providing comprehensive interpretable insights for further decision-making through human evaluation. This level of interpretability is not achievable by traditional approaches. To the best of our knowledge, we are the pioneers in directly using LLMs for the task of TSAD.

\bibliographystyle{ACM-Reference-Format}
\bibliography{sample-base}

\clearpage
\appendix
\section{Implementation details}
\subsection{Hyperparameters}
\textbf{GPT-4}: We use GPT-4-1106-preview as our baseline model for the main and ablation experiments. A grid search is performed over \(\alpha \in [0.8, 0.95, 0.99]\), \(\beta \in [0, 0.05, 0.15]\), with a temperature of 0.7.

\textbf{GPT-3.5}: We use GPT-3.5-turbo and perform a grid search over \(\alpha \in [0.8, 0.95, 0.99]\), \(\beta \in [0, 0.05, 0.15]\), with a temperature of 0.7.

\textbf{Llama}: We select the Llama-3-70B-instruct version as the comparison model and perform a grid search over \(\alpha \in [0.8, 0.95, 0.99]\), \(\beta \in [0, 0.05, 0.15]\), with a temperature of 1.0.

\section{Visualization Results}
\begin{figure*}[htbp]
\centering
\includegraphics[scale=1.3]{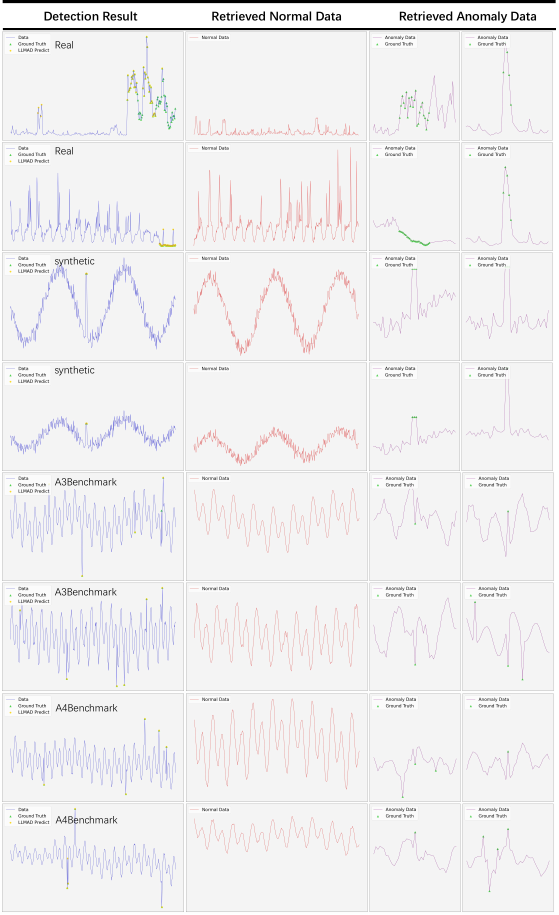}
\caption{Visualization of anomaly detection results and retrieved ICL data for the Yahoo datase.}
\label{fig:yahoo_ful}
\end{figure*}

Figure \ref{fig:yahoo_ful} presents visualizations of \name's predictions (a GPT-4-based model) and retrieval results on the Yahoo datasets for a subset of each time series set.



\section{Interpretability Evaluation Criteria}
\label{interpret_def}
To comprehensively evaluate the explanation text generating by \name, we adopt usefulness and readability as the primary criteria. These criteria are instrumental in assessing how the explanations generated by \name facilitate a better understanding and actionable insights into detected anomalies. Specifically, 
\begin{enumerate}[leftmargin=*]
    \item \textbf{Usefulness}: This criterion evaluates the extent to which the model's explanations contribute to the accurate and efficient resolution of anomalies. Explanations are scored on a scale from 1 (incorrect and irrelevant) to 5 (correct, detailed, and clear), with higher scores indicating that the explanation effectively assists in diagnosing and resolving anomalies, as well as accurately determining the urgency level.
    \item \textbf{Readability}: This criterion assesses the ease with which users can comprehend the explanations, focusing on their human-readability. This factor is essential for ensuring that the insights are both accessible and actionable. It is measured on a scale from 1 (difficult to understand) to 3 (high quality), with higher scores signifying enhanced clarity and ease of comprehension.
\end{enumerate}
The aforementioned criteria, \textbf{usefulness} and \textbf{readability}, are widely employed for LLM evaluation due to their significance in assessing the practicality and accessibility of generated explanations. A high usefulness score suggests that the explanations are accurate and beneficial for decision-making processes. Likewise, a high readability score indicates that the information is user-friendly, thereby improving operational efficiency by facilitating a better comprehension of the explanations and underlying reasoning.

\begin{table}[]
\centering
\caption{The criteria for the evaluation of usefulness scores.}
\label{tab:usefulness_def}
\resizebox{\columnwidth}{!}{%
\begin{tabular}{@{}ll@{}}
\toprule
\multicolumn{1}{c}{\textbf{Score type}} &
  \multicolumn{1}{c}{\textbf{Description of usefulness}} \\ \midrule
1 Irrelevant &
  \begin{tabular}[c]{@{}l@{}}Explanations don't match data or contain logical   \\ inconsistencies. Inaccurate alarm level.\end{tabular} \\ \midrule
2 No Useful Explanation &
  Merely restates rules without specifying alarm level. \\ \midrule
3 Basic Support &
  \begin{tabular}[c]{@{}l@{}}Supports anomaly detection with partial explanations;\\ broadly categorized alarm level.\end{tabular} \\ \midrule
4 Accurate and Relevant &
  \begin{tabular}[c]{@{}l@{}}Offers precise explanations aiding analysis and false \\ positive reduction. Detects all anomalies, yet lacks \\ detailed clarity; alarm level accurately set.\end{tabular} \\ \midrule
5 Detailed and Clear &
  \begin{tabular}[c]{@{}l@{}}Comprehensive aid in root cause identification. \\ Detects all anomalies,  guiding root cause localization; \\ alarm level precisely calibrated.\end{tabular} \\ \bottomrule
\end{tabular}%
}
\end{table}

\begin{table}[]
\centering
\caption{The criteria for the evaluation of readability scores.}
\label{tab:readability_def}
\resizebox{\columnwidth}{!}{%
\begin{tabular}{@{}ll@{}}
\toprule
\multicolumn{1}{c}{\textbf{Score type}} & \multicolumn{1}{c}{\textbf{Description of readability}}                                                                        \\ \midrule
1 Hard to Understand               & Confusing text with unclear transitions.                                                                                       \\ \midrule
2 Moderately Readable              & \begin{tabular}[c]{@{}l@{}}Generally clear, but awkward phrasing or \\ complex structures may hinder flow.\end{tabular}        \\ \midrule
3 High Quality                     & \begin{tabular}[c]{@{}l@{}}Clear, concise, and technically well-expressed; \\ suitable for professional contexts.\end{tabular} \\ \bottomrule
\end{tabular}%
}
\end{table}

\section{Detailed Prompt}
Different datasets have varying definitions of anomalies. Based on the feature of each dataset, we made specific adjustments to the base prompt in Figure \ref{fig:prompt}.
\subsection{WSD datasets}
The TSAD prompt for the WSD dataset is shown in Table \ref{detail_prompt_WSD}.

\begin{table*}[h!]
\centering
\caption{The prompt for WSD dataset.}
\label{detail_prompt_WSD}
\begin{tabular}{p{0.95\linewidth}}
\hline
\#\#\textbf{Instructions} \\
Determine if there are any anomalies in the provided AIOPS flow data sequence. \\
\#\#\textbf{Following Rules:} \\
1. A data point is considered an anomaly if it is part of a sequence of at least one consecutive anomalous points or continues to plummet or surge abruptly. \\
2. A data point is considered an anomaly if it is identified as a continuous low/high value anomaly if it remains below/above a predefined normal threshold for a prolonged duration, deviating from the anticipated norm. \\
3. Given that the vast majority of data points are expected to be no anomaly, Anomalies are exceedingly rare and should only be identified with absolute certainty. \\
4. Normal data may exhibit volatility, which should not be mistaken for anomalies. \\
5. Mislabeling normal data as an anomaly can lead to catastrophic failures. Exercise extreme caution. False positives are unacceptable. \\
6. If do not have 100 percent confidence that data is an anomaly, do not flag it as an anomaly. \\
7. The output of anomaly intervals needs to be accurately located and should not be excessively long. \\
8. The number of abnormal intervals within a detection range can not exceed 3. \\
9. anomaly\_type should be one of the following: \\
  - PersistentLevelShiftUp: The data shifts to a higher value and maintains that level consistently, do not return to the original baseline. like \texttt{1 2 1 2 1 2 *500* *480* *510* *500* *500*} \\
  - PersistentLevelShiftDown: The data shifts to a lower value and maintains that level consistently, do not return to the original baseline. like \texttt{1 2 1 2 *-100* *-102* *-104* *-110* *-110*} \\
  - TransientLevelShiftUp: The data temporarily shifts to a higher value and then returning to the original baseline, the anomaly maintains for at least 5 data points. like \texttt{1 2 1 2 1 2 *500* *500* *499* *510* *500* 1 2 1 2} \\
  - TransientLevelShiftDown: The data temporarily shifts to a lower value and then returning to the original baseline, the anomaly maintains for at least 5 data points. like \texttt{1 2 1 2 *-100* *-102* *-104* *-110* *-100* 1 2 1 2} \\
  - SingleSpike: A brief, sharp rise in data value followed by an immediate return to the baseline. like \texttt{1 2 1 2 1 2 *200* *500* 1 2} \\
  - SingleDip: A brief, sharp drop in data value followed by an immediate return to the baseline. like \texttt{1 2 1 2 *-500* *-200* 1 2 1 2} \\
  - MultipleSpikes: Several brief, sharp rises in data value, each followed by a return to the baseline. like \texttt{1 2 *500* 3 2 *510* *200* 1 2 *480* 1 2} \\
  - MultipleDips: Several brief, sharp drops in data value, each followed by a return to the baseline. like \texttt{1 2 *-100* 3 2 *-110* *-200* 1 2 *-120* 1 2} \\
10. alarm\_level should be one of the following: \\
  - Urgent/Error: This category is for values that represent a severe risk, potentially causing immediate damage or harm across all event types whether increases, decreases, spikes, dips, or multiple occurrences. \\
  - Important: Allocated for moderate value changes (both increases and decreases) that could escalate to future problems or system stress but are not immediately hazardous. This also covers upward transient level shifts that concern system longevity and potential failure indications from downward shifts. \\
  - Warning: Used for noticeable deviations from the norm that are not yet critical but merit close monitoring. This includes single spikes and dips that are moderate in nature, as well as multiple non-critical spikes and level shifts that are significant but not yet dangerous. \\
11. The briefExplanation must comprise a explicit three-step analysis results utilizing precise data (do not only repeat the rule): \\
  - Step 1: Assess the overall trend to ascertain if it aligns with expected patterns, thereby identifying any overarching anomalies. \\
  - Step 2: Determine if there is any local data segment with any continuous low or high values compared to the normal data sequence. \\
  - Step 3: Reassess the identified points to confirm their anomalous nature, given the rarity of true anomalies. \\
12. Provide responses in a strict JSON format suitable for direct parsing, without any additional textual commentary. \\
\#\#\textbf{Response Format} \\
\texttt{\{ "briefExplanation": \{"step1\_global": analysis reason, "step2\_local": analysis reason, "step3\_reassess": analysis reason\}, "is\_anomaly": false/true, "anomalies": []/[index1, index2, index3, ...], "reason\_for\_anomaly\_type": "no"/"reason for anomaly type", "anomaly\_type": "no"/"classification of main anomaly",(only one) "reason\_for\_alarm\_level": "no"/"reason for alarm level", "alarm\_level": "no"/"Urgent/Error"/"Important"/"Warning",(only one) \} } \\
\#\#\textbf{Data} \\
Please analyze the latest data with the highest level of diligence and caution: \\
- Historical normal data sequence: \texttt{\{normal\_data\}} \\
- Historical anomaly data sequence(*XXX* is anomaly point), \texttt{\{anomaly\_data\}} \\
- The latest \texttt{\{data\_len\}} data points for evaluation: \texttt{\{data\}} \\
\hline
\end{tabular}
\end{table*}

\subsection{KPI datasets}
The TSAD prompt for the KPI dataset is shown in Table \ref{detail_prompt_KPI}.

\begin{table*}[h!]
\centering
\caption{The prompt for KPI dataset.}
\label{detail_prompt_KPI}
\begin{tabular}{p{0.95\linewidth}}
    \hline
\#\#\textbf{Instructions} \\
Determine if there are any anomalies in the provided AIOPS flow data sequence. \\
\#\#\textbf{Following Rules:} \\
1. A data point is considered an anomaly if it is part of a sequence of at least one consecutive anomalous points or continues to plummet or surge abruptly. \\
2. Given that the vast majority of data points are expected to be \textbf{no anomaly}, Anomalies are exceedingly rare and should only be identified with absolute certainty. \\
3. Normal data may exhibit volatility, which should not be mistaken for anomalies. \\
4. Mislabeling normal data as an anomaly can lead to catastrophic failures. Exercise extreme caution. False positives are unacceptable. \\
5. If do not have 100 percent confidence that data is an anomaly, do not flag it as an anomaly. \\
6. The output of anomaly intervals needs to be accurately located and should not be excessively long. \\
7. anomaly\_type should be one of the following: \\
  - PersistentLevelShiftUp: The data shifts to a higher value and maintains that level consistently, do not return to the original baseline. like \texttt{1 2 1 2 1 2 *500* *480* *510* *500* *500*} \\
  - PersistentLevelShiftDown: The data shifts to a lower value and maintains that level consistently, do not return to the original baseline. like \texttt{1 2 1 2 *-100* *-102* *-104* *-110* *-110*} \\
  - TransientLevelShiftUp: The data temporarily shifts to a higher value and then returning to the original baseline, the anomaly maintains for at least 5 data points and return to baseline like \texttt{1 2 1 2 1 2 *500* *500* *499* *510* *500* 1 2 1 2} \\
  - TransientLevelShiftDown: The data temporarily shifts to a lower value and then returning to the original baseline, the anomaly maintains for at least 5 data points return to baseline like \texttt{1 2 1 2 *-100* *-102* *-104* *-110* *-100* 1 2 1 2} \\
  - SingleSpike: A brief, sharp rise in data value followed by an immediate return to the baseline. like \texttt{1 2 1 2 1 2 *200* *500* 1 2} \\
  - SingleDip: A brief, sharp drop in data value followed by an immediate return to the baseline. like \texttt{1 2 1 2 *-500* *-200* 1 2 1 2} \\
  - MultipleSpikes: Several brief, sharp rises in data value, each followed by a return to the baseline. like \texttt{1 2 *500* 3 2 *510* *200* 1 2 *480* 1 2} \\
  - MultipleDips: Several brief, sharp drops in data value, each followed by a return to the baseline. like \texttt{1 2 *-100* 3 2 *-110* *-200* 1 2 *-120* 1 2} \\
8. alarm\_level should be one of the following: \\
  - Urgent/Error: This category is for values that represent a severe risk, potentially causing immediate damage or harm across all event types whether increases, decreases, spikes, dips, or multiple occurrences. \\
  - Important: Allocated for moderate value changes (both increases and decreases) that could escalate to future problems or system stress but are not immediately hazardous. This also covers upward transient level shifts that concern system longevity and potential failure indications from downward shifts. \\
  - Warning: Used for noticeable deviations from the norm that are not yet critical but merit close monitoring. This includes single spikes and dips that are moderate in nature, as well as multiple non-critical spikes and level shifts that are significant but not yet dangerous. \\
9. The briefExplanation must comprise a explicit three-step analysis utilizing precise data (do not only repeat the rule): \\
  - Step 1: Assess the overall trend to ascertain if it aligns with expected patterns, thereby identifying any overarching anomalies. \\
  - Step 2: Examine the local data segments to detect any specific deviations or anomalies. \\
  - Step 3: Reassess the identified points to confirm their anomalous nature, given the rarity of true anomalies. This step ensures that the detected points are not merely normal fluctuations or seasonal variations. \\
10. Provide responses in a strict JSON format suitable for direct parsing, without any additional textual commentary. \\
\#\#\textbf{Response Format} \\
\texttt{\{ "briefExplanation": \{"step1\_global": analysis reason, "step2\_local": analysis reason, "step3\_reassess": analysis reason\}, "is\_anomaly": false/true, "anomalies": []/[index1, index2, index3, ...], "reason\_for\_anomaly\_type": "no"/"reason for anomaly type", "anomaly\_type": "no"/"classification of main anomaly",(only one) "reason\_for\_alarm\_level": "no"/"reason for alarm level", "alarm\_level": "no"/"Urgent/Error"/"Important"/"Warning",(only one) \} } \\
\#\#\textbf{Data} \\
Please analyze the latest data with the highest level of diligence and caution: \\
- Historical normal data sequence: \texttt{\{normal\_data\}} \\
- Historical anomaly data sequence(*XXX* is anomaly point), \texttt{\{anomaly\_data\}} \\
- The latest \texttt{\{data\_len\}} data points for evaluation: \texttt{\{data\}} \\
\hline
\end{tabular}
\end{table*}

\subsection{Yahoo datasets}
Given that the Yahoo dataset comprises multiple sub-datasets, we developed two prompts. The first prompt is tailored for the Yahoo real dataset and is analogous to the prompts used for KPI and WSD, as detailed in Table \ref{detail_prompt_yahoo_s}. The second prompt is designed for the A3Benchmark, A4Benchmark, and synthetic sub-datasets. In these sub-datasets, most anomalies are local; therefore, we focus exclusively on detecting single spikes and single dips, omitting the global step from AnoCoT, as illustrated in Table \ref{detail_prompt_yahoo_r}.

\begin{table*}[h!]
\centering
\caption{The prompt for Yahoo A3Benchmark, A4Benchmark, and Synthetic sub-datasets.}
\label{detail_prompt_yahoo_s}
\begin{tabular}{>{\raggedright\arraybackslash}p{0.95\textwidth}}
\hline
\#\#\textbf{Instructions} \\
Determine if there are any anomalies in the provided AIOPS flow data sequence. \\
\#\#\textbf{Following Rules:} \\
1. A data point is considered an anomaly if it is part of a sequence of at least one consecutive anomalous points or continues to plummet or surge abruptly. \\
2. Typically, anomalies are outliers such as spikes and dips, which are often isolated points. Be aware that there may be multiple anomalies present; you should identify all possible anomalous data points. \\
3. Given that the vast majority of data points are expected to be no anomaly, Anomalies are exceedingly rare and should only be identified with absolute certainty. \\
4. Mislabeling normal data as an anomaly can lead to catastrophic failures. Exercise extreme caution. False positives are unacceptable. \\
5. If do not have 100 percent confidence that data is an anomaly, do not flag it as an anomaly. \\
6. The output of anomaly intervals needs to be accurately located and should not be excessively long. \\
7. anomaly\_type should be one of the following: \\
- SingleSpike: A brief, sharp rise in data value followed by an immediate return to the baseline. like \texttt{1 2 1 2 1 2 *200* *500* 1 2} \\
- SingleDip: A brief, sharp drop in data value followed by an immediate return to the baseline. like \texttt{1 2 1 2 *-500* *-200* 1 2 1 2} \\
8. alarm\_level should be one of the following: \\
- Urgent/Error: This category is for values that represent a severe risk, potentially causing immediate damage or harm across all event types whether increases, decreases, spikes, dips, or multiple occurrences. \\
- Important: Allocated for moderate value changes (both increases and decreases) that could escalate to future problems or system stress but are not immediately hazardous. This also covers upward transient level shifts that concern system longevity and potential failure indications from downward shifts. \\
- Warning: Used for noticeable deviations from the norm that are not yet critical but merit close monitoring. This includes single spikes and dips that are moderate in nature, as well as multiple non-critical spikes and level shifts that are significant but not yet dangerous. \\
9. The briefExplanation must comprise a explicit two-step analysis \textit{results} utilizing precise data (do not only repeat the rule): \\
- Step 1: Examine the local data point to detect any specific deviations or anomalies. You should identify all possible anomalous data points. \\
- Step 2: Reassess the identified points to confirm their anomalous nature, given the rarity of true anomalies. \\
10. Provide responses in a strict JSON format suitable for direct parsing, without any additional textual commentary. \\
\#\#\textbf{Response Format} \\
\texttt{\{ "briefExplanation": \{"step1\_local": analysis reason, "step2\_reasses": analysis reason\}, "is\_anomaly": false/true, "anomalies": []/[index1, index2, index3, ...], "reason\_for\_anomaly\_type": "no"/"reason for anomaly type", "anomaly\_type": "no"/"classification of main anomaly", "reason\_for\_alarm\_level": "no"/"reason for alarm level", "alarm\_level": "no"/"Urgent/Error"/"Important"/"Warning" \} } \\
\#\#\textbf{Data} \\
Please analyze the latest data with the highest level of diligence and caution: \\
- Historical normal data sequence: \texttt{\{normal\_data\}} \\
- Historical anomaly data sequence(*XXX* is anomaly point), \texttt{\{anomaly\_data\}} \\
- The latest \texttt{\{data\_len\}} data points for evaluation: \texttt{\{data\}} \\
\hline
\end{tabular}
\end{table*}

\begin{table*}[h!]
\centering
\caption{The prompt for Yahoo real subset}
\label{detail_prompt_yahoo_r}
\begin{tabular}{>{\raggedright\arraybackslash}p{0.95\textwidth}}
\hline
\#\#\textbf{Instructions:} \\
Determine if there are any anomalies in the provided AIOPS flow data sequence. \\
\#\#\textbf{Following Rules:} \\
1. A data point is considered an anomaly if it is part of a sequence of at least one consecutive anomalous points or continues to plummet or surge abruptly. \\
2. A data point is considered an anomaly if it is identified as a continuous low/high value anomaly if it remains below/above a predefined normal threshold for a prolonged duration, deviating from the anticipated norm. \\
3. Given that the vast majority of data points are expected to be no anomaly, Anomalies are exceedingly rare and should only be identified with absolute certainty. \\
4. Normal data may exhibit volatility, which should not be mistaken for anomalies. \\
5. Mislabeling normal data as an anomaly can lead to catastrophic failures. Exercise extreme caution. False positives are unacceptable. \\
6. If do not have high percent confidence that data is an anomaly, do not flag it as an anomaly. \\
7. The output of anomaly intervals needs to be accurately located and should not be excessively long. \\
8. The number of abnormal intervals within a detection range can not exceed 3. \\
9. anomaly\_type should be one of the following: \\
  - PersistentLevelShiftUp The data shifts to a higher value and maintains that level consistently, do not return to the original baseline. like \texttt{1 2 1 2 1 2 *500* *480* *510* *500* *500*} \\
  - PersistentLevelShiftDown The data shifts to a lower value and maintains that level consistently, do not return to the original baseline. like \texttt{1 2 1 2 *-100* *-102* *-104* *-110* *-110*} \\
  - TransientLevelShiftUp The data temporarily shifts to a higher value and then returning to the original baseline, the anomaly maintains for at least 5 data points. like \texttt{1 2 1 2 1 2 *500* *500* *499* *510* *500* 1 2 1 2} \\
  - TransientLevelShiftDown The data temporarily shifts to a lower value and then returning to the original baseline, the anomaly maintains for at least 5 data points. like \texttt{1 2 1 2 *-100* *-102* *-104* *-110* *-100* 1 2 1 2} \\
  SingleSpike A brief, sharp rise in data value followed by an immediate return to the baseline. like \texttt{1 2 1 2 1 2 *200* *500* 1 2} \\
  SingleDip A brief, sharp drop in data value followed by an immediate return to the baseline. like \texttt{1 2 1 2 *-500* *-200* 1 2 1 2} \\
  MultipleSpikes Several brief, sharp rises in data value, each followed by a return to the baseline. like \texttt{1 2 *500* 3 2 *510* *200* 1 2 *480* 1 2} \\
  - MultipleDips Several brief, sharp drops in data value, each followed by a return to the baseline. like \texttt{1 2 *-100* 3 2 *-110* *-200* 1 2 *-120* 1 2} \\
10. alarm\_level should be one of the following: \\
  - Urgent/Error This category is for values that represent a severe risk, potentially causing immediate damage or harm across all event types whether increases, decreases, spikes, dips, or multiple occurrences. \\
  - Important Allocated for moderate value changes (both increases and decreases) that could escalate to future problems or system stress but are not immediately hazardous. This also covers upward transient level shifts that concern system longevity and potential failure indications from downward shifts. \\
  - Warning Used for noticeable deviations from the norm that are not yet critical but merit close monitoring. This includes single spikes and dips that are moderate in nature, as well as multiple non-critical spikes and level shifts that are significant but not yet dangerous. \\
11. The briefExplanation must comprise a explicit three-step analysis \textit{results} utilizing precise data (do not only repeat the rule): \\
  - Step 1: Assess the overall trend to ascertain if it aligns with expected patterns, thereby identifying any overarching anomalies. \\
  - Step 2: Determine if there is any local data segment with any continuous low or high values compared to the normal data sequence. \\
  - Step 3: Reassess the identified points to confirm their anomalous nature, given the rarity of true anomalies. \\
12. Provide responses in a strict JSON format suitable for direct parsing, without any additional textual commentary. \\
\#\#\textbf{Response Format:} \\
\texttt{\{ "briefExplanation": \{"step1\_global": analysis reason, "step2\_local": analysis reason, "step3\_reassess": analysis reason\}, "is\_anomaly": false/true, "anomalies": []/[index1, index2, index3, ...], "reason\_for\_anomaly\_type": "no"/"reason for anomaly type", "anomaly\_type": "no"/"classification of main anomaly",(only one) "reason\_for\_alarm\_level": "no"/"reason for alarm level", "alarm\_level": "no"/"Urgent/Error"/"Important"/"Warning",(only one) \} } \\
\#\#\textbf{Data:} \\
Please analyze the latest data with the highest level of diligence and caution: \\
- Historical normal data sequence: \texttt{\{normal\_data\}} \\
- Historical anomaly data sequence(*XXX* is anomaly point), \texttt{\{anomaly\_data\}} \\
- The latest \texttt{\{data\_len\}} data points for evaluation: \texttt{\{data\}} \\
\hline
\end{tabular}
\end{table*}

\end{document}